\documentclass{article}

\pdfoutput=1

\usepackage[round,comma]{natbib}
\usepackage{geometry}
\usepackage{amssymb}
\usepackage{amsmath}
\allowdisplaybreaks[1] 
\usepackage{bbold} 

\usepackage[pdftex]{graphicx}
\DeclareGraphicsExtensions{.pdf}
\usepackage{flafter} 

\usepackage{fancyhdr} 
\usepackage{ifthen} 
\usepackage{layout} 
\usepackage{xspace}
\usepackage{afterpage}
\usepackage{url}
\usepackage{textcomp} 

\usepackage{wasysym} 

\usepackage[latin1]{inputenc} 

\def\ba#1\ea{\begin{align*}#1\end{align*}} 
\def\banum#1\eanum{\begin{align}#1\end{align}} 

\def\bi#1\ei{\begin{itemize}#1\end{itemize}} 

\newcommand{\ulesqed}{\hfill\smiley}

\ifx\proof\undefined
\newenvironment{proof}{\par\noindent{\em Proof. }}{\ulesqed\\[2mm]}
\else
\fi

\newcommand{\Nat}{\mathbb{N}} 

\DeclareMathOperator{\rreg}{R_{reg}}
\DeclareMathOperator{\remp}{R_{emp}}

\DeclareMathOperator{\Remp}{R_{emp}}

\renewcommand{\phi}{\ensuremath{\varphi}}

\newcommand{\condon}{\; \big| \;}
\let\Pr\undefined 
\DeclareMathOperator{\Pr}{Pr}

\wasyfamily
\DeclareFontShape{U}{wasy}{b}{n}{ <-10> ssub * wasy/m/n
<10> <10.95> <12> <14.4> <17.28> <20.74> <24.88>wasyb10 }{}

\DeclareFontShape{U}{wasy}{m}{n}{ <5> <6> <7> <8> <9> gen * wasy
<10> <10.95> <12> <14.4> <17.28> <20.74> <24.88> <35> <40> <50> <60> wasy10  }{}

\usepackage{amssymb}
\usepackage{amsbsy}
\usepackage{ifthen}
\usepackage{eucal}
\usepackage{theorem}

\newtheorem{theorem}{Theorem}
\newtheorem{definition}[theorem]{Definition}

\newtheorem{lemma}[theorem]{Lemma}

\newcommand{\commentout}[1]{}

\renewcommand{\Pr}{P}

\newcommand{\R}{\mathbb{R}}    

\newcommand{\Fcal}{\mathcal{F}} 
\newcommand{\Rcal}{\mathcal{R}} 
\newcommand{\Ncal}{\mathcal{N}} 
\newcommand{\Xcal}{\mathcal{X}} 
\newcommand{\Ycal}{\mathcal{Y}} 
\newcommand{\Pcal}{\mathcal{P}} 

\newcommand{\Eb}{E}    

\newcommand{\eps}{\varepsilon}              

\renewcommand{\k}{k}

\newcommand{\sgn}{\mathop{\mathrm{sgn}\,}}
\newcommand{\argmin}{\mathop{\mathrm{argmin}\,}}
\newcommand{\argmax}{\mathop{\mathrm{argmax}\,}}

\newcommand{\emp}{\mathrm{emp}}      

\newcommand{\fig}[1]{Figure~\protect\ref{#1}}

\newcommand{\eq}[1]{(\protect\ref{#1})}

  \mathchardef\ordinarycolon\mathcode`\:
  \mathcode`\:=\string"8000
  \begingroup \catcode`\:=\active
    \gdef:{\mathrel{\mathop\ordinarycolon}}
  \endgroup

\newcommand{\note}[1]{\hspace{0.3cm}\text{\em (#1) }}
\newcommand{\VC}{\text{VC}}
\newcommand{\nn}{\text{NN}}
\newcommand{\knn}{\text{kNN}}

\begin{document}

\title{Statistical Learning Theory: Models, Concepts, and Results}

\author{Ulrike von Luxburg\\
Max Planck Institute for Biological Cybernetics\\
T{\"u}bingen, Germany \\
\url{ulrike.luxburg@tuebingen.mpg.de}  
\and 
Bernhard Sch\"olkopf\\
Max Planck Institute for Biological Cybernetics\\
T{\"u}bingen, Germany \\
\url{bernhard.schoelkopf@tuebingen.mpg.de}  }
\date{September 2008}
\maketitle

\section{Introduction } \label{sec-intro}

Statistical learning theory %
provides the
theoretical basis for many of today's machine learning algorithms and
is arguably one of the most beautifully developed branches of
artificial intelligence in general. It originated  in
Russia in the 1960s and gained wide popularity in the 1990s 
following the development of the so-called {\em Support Vector
  Machine (SVM)}, which has become a standard tool for pattern recognition in
a variety of domains ranging from computer vision to computational
biology. Providing the basis of new learning algorithms, however, was
not the only motivation for developing statistical learning theory. 
It was just as much a philosophical one, attempting to answer the question
of what it is that allows us to draw valid conclusions from empirical data.\\

In this article we attempt to give a gentle, non-technical overview
over the key ideas and insights of statistical learning theory. We do
not assume that the reader has a deep background in mathematics,
statistics, or computer science. Given the nature of the subject
matter, however, some familiarity with mathematical concepts and
notations and some intuitive understanding of basic probability is
required.  There exist many excellent references to more technical
surveys of the mathematics of statistical learning theory: the
monographs by one of the founders of statistical learning theory
(\citealp{Vapnik95}, \citealp{Vapnik98}), a brief overview over
statistical learning theory in
Section~5 of \citet{SchSmo02}, more technical overview papers such as
\citet{BouBouLug03}, \citet{Mendelson03}, \citet{BouBouLug05}, 
\citet{HerWil02}, and the
monograph  \citet{DevGyoLug96}. \\

\section{The standard framework of statistical learning theory} \label{sec-framework}
\subsection{Background} \label{subsec-background}

In our context, learning refers to the process of inferring
general rules by observing examples. Many living organisms show some
ability to learn. For instance, children can learn 
what ``a car'' is, just by being shown examples of objects that
are cars and objects that are not cars. They do not need to be 
told any rules about what is it that makes an object a
car, they can simply learn the concept ``car'' by observing examples. \\

The field of machine learning does not study the process of learning
in living organisms, but instead studies the process of learning in
the abstract. The question is how a machine, a computer, can ``learn''
specific tasks by following specified learning algorithms. To this
end, the machine is shown particular examples of a specific task. Its
goal is then to infer a general rule which can both explain the
examples it has seen already and which can generalize to previously
unseen, new examples. Machine learning has roots in artificial
intelligence, statistics, and computer science, but by now has
established itself as a scientific discipline in its own right. As
opposed to artificial intelligence, it does not try to explain or
generate ``intelligent behavior'', its goal is more modest: it just
wants to discover mechanisms by which very specific tasks can be
``learned'' by a computer. Once put into a formal framework, many of
the problems studied in machine learning sound familiar from
statistics or physics: regression, classification, clustering, and so
on. However, machine learning looks at those problems with a different
focus: the one of inductive inference and generalization ability.
\\

The most well-studied problem in machine learning is the problem of
classification. Here we deal with two kind of spaces: the input space
$\Xcal$ (also called space of instances) and the output space (label
space) $\Ycal$.  For example, if the task is to classify certain
objects into a given, finite set of categories such as ``car'',
``chair'', ``cow'', then $\Xcal$ consists of the space of all possible
objects (instances) in a certain, fixed representation, while $\Ycal$
is the space of all available categories. In order to learn, an
algorithm is given some training examples $(X_1, Y_1), ..., (X_n,
Y_n)$, that is pairs of objects with the corresponding category label.
The goal is then to find a mapping $f: \Xcal \to \Ycal$ which makes
``as few errors as possible''. That is, among all the elements in
$\Xcal$, the number of objects which are assigned to the wrong
category is as small as
possible. The mapping $f: \Xcal \to \Ycal$ is called a classifier. \\

In general, we distinguish between two types of learning problems:
supervised ones and unsupervised ones. Classification is an example
of a supervised learning problem: the training examples consist both
of instances $X_i$ and of the correct labels $Y_i$ on those instances. The goal is
to find a functional relationship between instances and outputs. This
setting is called supervised because at least on the training
examples, the learner can evaluate whether an answer is correct,
that is the learner is being supervised.  Contrary to this, the training data
in the unsupervised setting only consists of instances $X_i$, without
any further 
information about what kind of output is expected on those instances.  In this setting,
the question of learning is more about discovering some ``structure''
on the underlying space of instances. A standard example of such a
setting is clustering. Given some input points $X_1, ..., X_n$, the
learner is requested to construct ``meaningful groups'' among the
instances. For example, an online retailer might want to cluster his
customers based on shopping profiles. He collects all kinds of
potentially meaningful information about his customers (this will lead
to the input $X_i$ for each customer) and then wants
to discover groups of customers with similar behavior. As opposed to
classification, however, it is not specified beforehand which customer
should belong to which group -- it is the task of the clustering
algorithm to work that out.\\

Statistical learning theory (SLT) is a theoretical branch of machine
learning and attempts to lay the mathematical foundations for the
field. The questions asked by SLT are fundamental: 

\begin{itemize}
\item Which learning tasks can be performed by computers in general (positive and negative results)? 
\item What kind of assumptions do we have to make such that machine learning
  can be successful? 
\item What are the key properties a learning algorithm needs to satisfy in order to be successful? 
\item Which performance guarantees can we give on the results of certain learning algorithms? 
\end{itemize}

To answer those questions, SLT builds on a certain mathematical
framework, which we are now going to introduce.  In the following, we
will focus on the case of supervised learning, more particular on the
case of binary classification. We made this choice because the theory for
supervised learning, in particular classification, is rather mature, while the theory for many 
branches of unsupervised learning is still in its infancy.\\

\subsection{The formal setup}  \label{subsec-formal}

In supervised learning, we deal with an {\em input space (space of
  instances, space of objects) $\Xcal$} and an {\em output space
  (label space)} $\Ycal$.  In the case of binary classification, we
identify the label space with the set $\{-1, +1\}$. That is, each
object can belong to one out of two classes, and by convention we
denote those classes by $-1$ and $1$. The question of learning is
reduced to the question of estimating a functional relationship of the
form $f: \Xcal \to \Ycal$, that is a relationship between input and
output. Such a mapping $f$ is called a {\em classifier}.  In order to do
this, we get access to some {\em training points (training
  examples, training data)} $(X_1, Y_1), ..., (X_n, Y_n) \in \Xcal
\times \Ycal$. A {\em classification algorithm (classification rule)}
is a procedure that takes the training data as input and outputs a
classifier $f$.  We do
not make specific assumptions on the spaces $\Xcal$ or $\Ycal$, but we
do make an assumption on the mechanism which generates those training
points. Namely, we assume that there exists a {\em joint probability
  distribution $P$} on $\Xcal \times \Ycal$, and the training examples
$(X_i, Y_i)$ are sampled independently from this distribution
$P$. This type of sampling is often denoted as iid sampling
(independent and identically distributed). There are a few important
facts to note here.

\begin{enumerate} 
\item {\em No assumptions on $P$}. In the standard setting of SLT we
  do not make any assumption on the probability distribution $P$: it
  can be any distribution on $\Xcal \times \Ycal$. In this sense,
  statistical learning theory works in an agnostic setting which is
  different from standard statistics, where one usually assumes
  that the probability distribution belongs to a certain family of
  distributions and the goal is to estimate the parameters of this
  distribution.
\item {\em Non-deterministic labels due to label noise or overlapping
    classes. } Note that $P$ is a probability distribution not only
  over the instances $\Xcal$, but also over the labels $\Ycal$. As a
  consequence, labels $Y_i$ in the data are not necessarily just a
  deterministic function of the objects $X_i$, but can be random
  themselves. There are two main reasons why this can be the case. The
  first reason is that the data generating process can be subject to
  label noise.  That is, it can happen that label $Y_i$ we get as a
  training label in the learning process is actually wrong. This is an
  important and realistic assumption. For example, to generate
  training data for email spam detection, humans are required to label
  emails by hand into classes ``spam'' and ``not spam''. All humans
  make mistakes from time to time. So it will happen that some emails
  accidentally get labeled as ``spam'' even though they are not spam,
  or vice versa. Of course, the hope is that such wrong labels only
  occur with a relatively small probability.  The second major reason
  which can lead to non-deterministic labels is the case of
  overlapping classes. As an example, consider the task of predicting
  the gender of a person based on their height. It is clear that a
  person of height 1.80 meters, say, could in principle be male or
  female, thus we cannot assign a
  unique label $Y$ to the input $X= 1.80$. \\

  For the purpose of learning, we will see that in the end it does not
  matter which of the reasons is the one leading to non-deterministic
  labels. It will turn out that the important quantity which covers
  both cases is the conditional  likelihood of the labels, namely the
  probability that the label $Y$ is 1, under the condition that the
  data point under consideration is the point $x$: 
\banum \label{eq-regression-fct}
\eta(x) := P( Y = 1 \condon X = x). 
\eanum
Note that we only consider the case $Y=1$ as the case $Y=-1$ can
simply be computed by $P(Y = -1 \condon X = x) = 1 - \eta(x)$. In the
case of small label noise, the conditional probability $\eta(x)$ is
either close to 1 or close to 0 (depending on whether the true label
is +1 or -1). For large label noise, on the other hand, the
probability $\eta(x)$ gets closer to 0.5 and learning becomes
more difficult. Similar reasoning applies to the case of overlapping
classes. In the example of predicting the gender of a person based on
height, the classes overlap quite strongly. For example, the
probability $P(Y = \text{"male"} \condon X = 1.70)$ might only be
0.6. That is, if we want to predict the gender of a person of height
1.70 we will, on average, make an error of at least 40\%. Both in
cases of label noise and of overlapping classes, learning becomes more
difficult the closer the function $\eta(x)$ comes to 0.5, and it
becomes unavoidable that a classifier makes a relatively large number
of errors.

\item {\em Independent sampling. } It is an important assumption of
  SLT that data points are sampled independently. This is a rather
  strong assumption, which is justified in many applications, but not in
  all of them. For example, consider the example of pattern
  recognition for hand written digits. Given some images of hand
  written digits, the task is to train a machine to automatically
  recognize new hand written digits. For this task, the training set
  usually consists of a large collection of digits written by many
  different people. Here it is safe to assume that those digits form
  an independent sample from the whole ``population'' of all hand written
  digits.  As an example where the independence
  assumption is heavily violated, consider the case of drug
  discovery. This is a field in pharmacy where people try to identify
  chemical compounds which might be helpful for designing new
  drugs. Machine learning is being used for this purpose: the training
  examples consist of chemical compounds $X_i$ with a label $Y_i$
  which indicates whether this compound is useful for drug design or
  not. It is expensive to find out whether a chemical compound possesses
certain properties that render it a suitable drug because this would
require running extensive lab experiments.  As a result, only rather few compounds $X_i$
  have known labels $Y_i$, and those compounds have been
  carefully selected in the first place. Here, we cannot assume that
  the $X_i$ are a representative sample drawn independently
  from some distribution of chemical compounds, as the labeled
  compounds are hand-selected according to some non-random process.\\

  Note that in some areas of machine learning, researchers try to
  relax the independence assumption. For example in active learning
  one deals with the situation where users can actively select the
  points they want to get labeled. Another case is time series
  prediction, where training instances are often generated from
  overlapping (and thus dependent) windows of a temporal sequence. We
  are not going to discuss those areas in this paper.

\item {\em The distribution $P$ is fixed. } In the standard setting of
  SLT, we do not have any ``time'' parameter. In particular, we do not
  assume any particular ordering of the training examples, and the
  underlying probability distribution does not change over time. This
  assumption would not be true if we wanted to argue about time
  series, for example.  Another situation that has recently attracted
  attention is the case where training and test distributions differ
  in certain aspects (e.g., under the heading of ``covariate shift'').

\item {\em The distribution $P$ is unknown at the time of learning. }
  It is important to recall that at the time of training the
  underlying distribution is not known. We will see below that if we
  knew $P$, then learning would be trivial as we could simply write
  down the best classifier by a given formula. Instead, we only have
  access to $P$ indirectly, by observing training
  examples. Intuitively this means that if we get enough training
  examples, we can ``estimate'' all important properties of $P$ pretty
  accurately, but we are still prone to errors. It is one of the big
  achievements of statistical learning theory to provide a
  framework to make theoretical statements about this error.

\end{enumerate}

As already mentioned above, the goal of supervised learning is to
learn a function $f: \Xcal \to \Ycal$. In order to achieve this, we
need to have some measure of ``how good'' a function $f$ is when used
as a classifier. To this end,  we introduce a {\em loss
  function}. This is a function $\ell$ 
which tells us the ``cost'' of classifying instance $X \in
\Xcal$ as $Y \in \Ycal$.  For example, the simplest loss function in
classification is the 0-1-loss or misclassification error: the loss of classifying $X$ by label
$f(X)$ is 0 if $f(X)$ is the correct label for $X$, and 1 otherwise: 

\ba
\ell(X, Y, f(X)) = 
\begin{cases}
1 & \text{ if } f(X) \neq Y\\
0 & \text{ otherwise.}
\end{cases}
\ea

In regression, where the output variables $Y$ take values that are real numbers rather
than class labels, a well-known loss function is the squared error
loss function $\ell(X, Y, f(X)) = (Y - f(X))^2$. The general convention
is that a loss of $0$ denotes perfect classification, and higher loss
values
represent worse classification performance. \\

While the loss function
measures the error of a function on some individual data point, the
{\em risk} of a function is the average loss over data points generated according to the underlying distribution $P$, 
\ba
R(f) := E ( \ell(X, Y, f(X)) ).
\ea
That is, the risk of a classifier $f$ is the expected loss
of the function $f$ at all points $X \in \Xcal$. 
Intuitively, this risk ``counts'' how many elements of the
instance space $\Xcal$ are misclassified by the function $f$. Of
course, a function $f$ is a better classifier than another function
$g$ if its risk is smaller, that is if $R(f) < R(g)$. To find a
good classifier $f$ we need to find one for which $R(f)$ is as small
as possible. The best classifier is the one with the smallest risk
value $R(f)$. \\

One aspect we have left open so far is what kind of functions $f$ to
consider. To formalize this, we consider some underlying {\em
  space $\Fcal$ of functions} which map $\Xcal$ to $\Ycal$. This is
the space of functions from which we want to choose our solution.  
At first glance, the most natural way would be to allow all possible
functions from $\Xcal$ to $\Ycal$ as classifier, that is to choose
$\Fcal_{all} = \{f: \Xcal \to \Ycal\}$. (We ignore issues about
measurability at this point; for readers familiar with measure theory,
note that one usually defines the space $\Fcal_{all}$ to be the space of
measurable functions between $\Xcal$ and $\Ycal$.) In this case, one
can formally write down what the optimal classifier should be. Given
the underlying probability
distribution $P$, this  classifier is defined as follows:\\

\banum \label{eq-fbayes}
f_{Bayes}(x) := 
\begin{cases}
  1 & \text{if } P(Y = 1 \condon X = x) \geq 0.5 \\
-1 & \text{ otherwise. }
\end{cases}
\eanum

This is the so-called ``{\em Bayes classifier}''.  Intuitively, what
it does is as follows. For each point in the space $\Xcal$, it looks
at the function $\eta(x) := P( Y = 1 \condon X = x)$ introduced in
Eq.~\eqref{eq-regression-fct}. If we assume that $P(Y=1 \condon X = x)
=1$, this means that the true label $Y$ of the point $X = x$ satisfies
$Y = 1$ with certainty (probability 1). Hence, an optimal classifier
should also take this value, that is it should choose $f(x) = 1$.  Now
assume that the classes slightly overlap, for example $P(Y=1
\condon X = x) = 0.9$. This still means that in an overwhelming
number of cases (in 90 \% of them), the label of object $x$ is $+1$,
thus this is what the classifier $f$ should choose. The same holds as
long as the overlap is so small that $\eta(x) \geq 0.5$. By choosing
$f(x) = 1$, the classifier $f$ will be correct in the majority of all
cases. Only when $\eta(x)$ goes below 0.5, the situation flips and the
optimal choice is to
choose $f = -1$. We will come back to this example in Section~\ref{sec-knn}.\\

In practice, it is impossible to directly compute the Bayes
classifier. The reason is that, as we explained above, the underlying
probability distribution is unknown to the learner. Hence, the Bayes
classifier cannot be computed, as we would need to evaluate the
conditional probabilities $P(Y=1 \condon X =x)$. With all those
definitions in mind, we can formulate the standard problem of
binary classification as follows: \\

{\em Given some training points $(X_1, Y_1), ..., (X_n, Y_n)$ which
  have been drawn independently from some unknown probability distribution $P$, and
  given some
  loss function $\ell$, how can we construct a function $f: \Xcal \to
  \Ycal$ which has risk $R(f)$ as close as possible to the risk of the
  Bayes classifier? \\
}

At this point, note that not only is it impossible to compute the
Bayes error, but also the risk of a function $f$ cannot be computed
without knowledge of $P$. All in all, it looks like a pretty
desperate situation: we have defined the goal of binary
classification (to minimize the risk of the classifier), and 
can even formally write down its solution (the Bayes classifier). But at the time of
training, we do not have access to the important quantities to compute
either of them. 
This is where SLT comes in. It provides a framework to analyze this
situation, to come up with solutions, and to provide guarantees on the
goodness of these solution. \\

\subsection{Generalization and consistency} \label{subsec-consistency}

There are a few more important notions we need to explain at this
point. The most important one is ``generalization''. 
Assume we are given some training set $(X_1, Y_1), ..., (X_n, Y_n)$,
and by some algorithm come up with a classifier $f_n$. Even though we
cannot compute the true underlying risk $R(f_n)$ of this classifier,
what we can do is to ``count'' the number of
mistakes on the training points. The resulting quantity also  has a name, it is called the {\em
  empirical risk} or the {\em training error}. Formally, for any
function $f$ it is defined
as

\ba
\remp(f) := \frac{1}{n} \sum_{i=1}^n \ell(X_i, Y_i, f(X_i)).
\ea

Usually, for a classifier $f_n$ learned on a particular training set,
the empirical risk $\remp(f_n)$ is relatively small -- otherwise, the
learning algorithm does not even seem to be able to explain the
training data. However, it is not clear whether a function $f_n$ which
makes few errors on the training set also makes few errors on the rest
of the space $\Xcal$, that is whether it has a small overall risk
$R(f_n)$. We say that a classifier $f_n$ {\em generalizes} well if the
difference $|R(f_n) - \remp(f_n)|$ is small. Note that with this
definition, good generalization performance does not necessarily mean
that a classifier has a small overall error $\remp$. It just means
that the empirical error $\remp(f_n)$ is a good estimate of the true
error $R(f_n)$. Particularly bad in practice is the situation where
$\remp(f_n)$ is much smaller than $R(f_n)$. In this case, using the
empirical risk as an estimator of the true risk would lead us to be
overly optimistic about the quality of our classifier.  \\

\begin{figure}[t]
    \centerline{\includegraphics[width=0.4\textwidth]{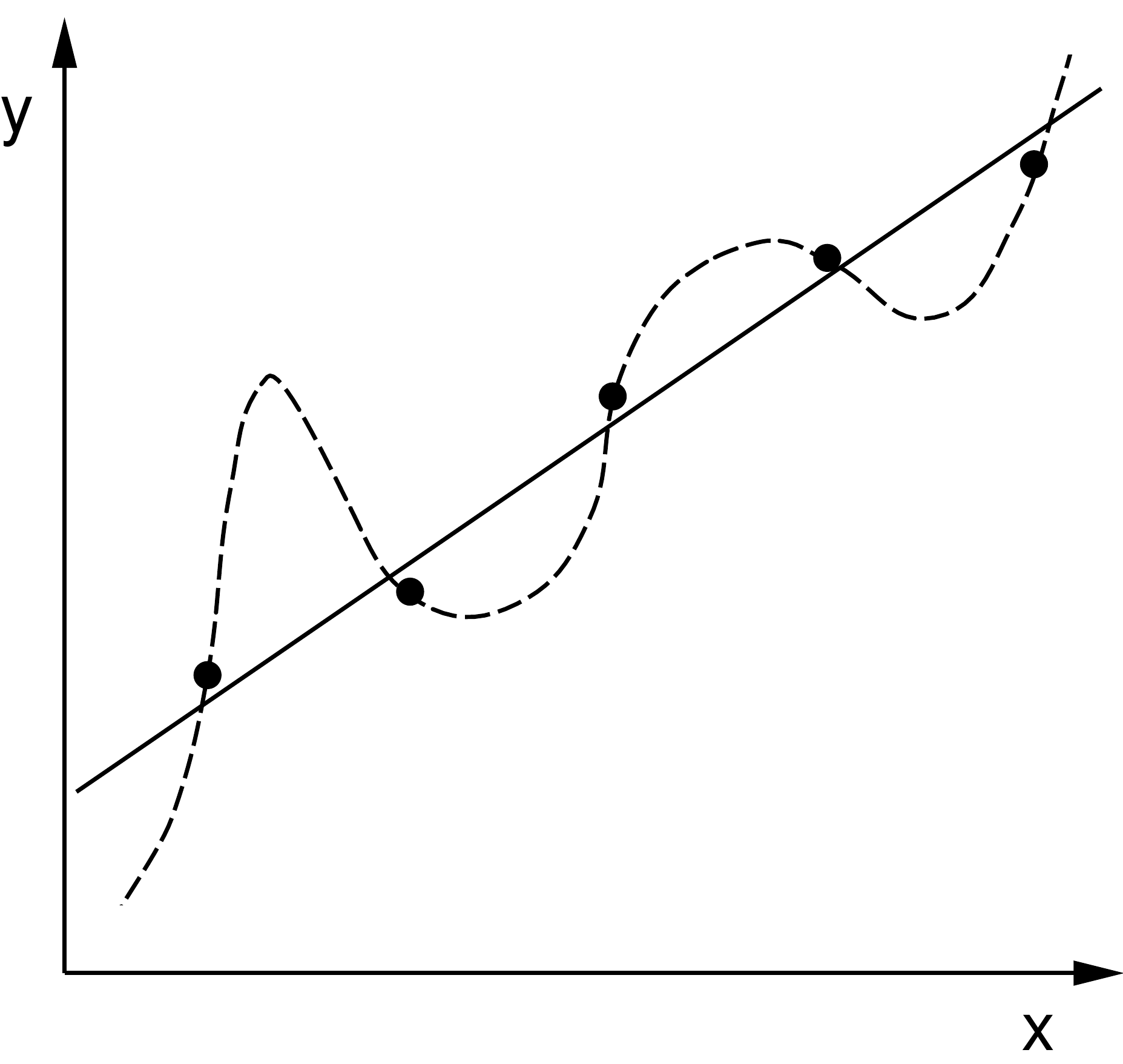}}
\caption{\em \label{fig:slt:fig1} Suppose we want to estimate a functional
  dependence from a set of examples (black dots). Which model is
  preferable? The complex model perfectly fits all data points,
  whereas the straight line exhibits residual errors. Statistical
  learning theory formalizes the role of the \emph{capacity} of the
  model class, and gives probabilistic guarantees for the validity of
  the inferred model (from \cite{SchSmo02}).}
\end{figure}

Consider the following regression example.
We are given empirical observations,
 $ (x_1,y_1),\dots,(x_m,y_m) \in \Xcal\times\Ycal $
 where for simplicity we take $\Xcal=\Ycal=\R$.  For example, the data
 could have been collected in a physical experiment where $X$ denotes
 the weight of an object, and $Y$ the force we need to pull this
 object over a rough surface.  \fig{fig:slt:fig1} shows a plot of such
 a dataset (indicated by the round points), along with two possible
 functional dependencies that could underlie the data. The dashed line
 $f_{dashed}$ represents a fairly complex model, and fits the training
 data perfectly, that is it has a training error of 0.  The straight line
 $f_{straight}$, on the other hand, does not completely ``explain''
 the training data, in the sense that there are some residual errors,
 leading to a small positive training error (for example, measured by
 the squared loss function).  But what about the true risks
 $R(f_{dashed})$ and $R(f_{straight})$? The problem is that we cannot
 compute this risk from the training data. Moreover, the functions
 $f_{dashed}$ and $f_{straight}$ have very different behavior.  For example, if the straight line was the
 true underlying function, then the dashed function $f_{dashed}$ would
 have a high true risk, as the ``distance'' between the true and
 the estimated function is very large. The same also holds the other
 way around. In both cases the true risk would be much higher than the
 empirical risk. \\

 This example points out an important choice we have to make. Do
 we prefer to fit the training data with a relatively ``complex''
 function, leading to a very small training error, or do we prefer to
 fit it with a ``simple'' function at the cost of a slightly higher
 training error? In the example above, a physicist measuring these
 data points would argue that it cannot be by chance that the
 measurements lie almost on a straight line and would much prefer to
 attribute the residuals to measurement error than to an erroneous
 model. But is it possible to {\em characterize} the way in which the
 straight line is simpler, and why this should imply that it is, in
 some sense, closer to an underlying true dependency?  What is the
 ``amount of increase in training error''
 we should be willing to tolerate for fitting a simpler model?\\

In one form or another, this issue has long occupied the minds of
researchers studying the problem of learning.  In classical
statistics, it has been studied as the \emph{bias-variance
  dilemma}. If we computed a linear fit for every data set that we
ever encountered, then every functional dependency we would ever
``discover'' would be linear. But this would not come from the data;
it would be a \emph{bias} imposed by us. If, on the other hand, we
fitted a polynomial of sufficiently high degree to any given data set,
we would always be able to fit the data perfectly, but the exact model
we came up with would be subject to large fluctuations, depending on
how accurate our measurements were in the first place --- the model
would suffer from a large \emph{variance}.  A related dichotomy is the
one between \emph{estimation error} and \emph{approximation error}. If
we use a small class of functions, then even the best possible
solution will poorly approximate the ``true'' dependency, while a
large class of functions will lead to a large statistical estimation
error.  We will discuss these dichotomies in more detail in Section \ref{subsec-bias-variance}. 
In the terminology of applied machine learning, the complex
explanation shows \emph{overfitting}, while an overly simple
explanation imposed by the learning machine design would lead to
\emph{underfitting}. \\

A concept which is closely related to generalization is the one of
{\em consistency}.  However, as  opposed to the notion of generalization
discussed above, consistency is not a property of an individual
function, but a property of a set of functions. As in classical
statistics, the notion of consistency aims to make a statement about
what happens in the
limit of infinitely many sample points. Intuitively, it seems
reasonable to request that a learning algorithm, when presented more
and more training examples, should eventually ``converge'' to an optimal
solution. \\

There exist two different types of consistency in the literature,
depending on the taste of the authors, and both of them are usually
just called ``consistency'' without any distinction. To introduce
these concepts, let us make the following notation. 
Given any particular classification algorithm, by $f_n$ we will denote
its outcome on a sample of $n$ training points. It is not important
how exactly the algorithm chooses this function. But note that any
algorithm chooses its functions from some particular function space
$\Fcal$. For some algorithms this space is given explicitly, for
others it only exists implicitly via the mechanism of the
algorithm. No matter how this space $\Fcal$ is defined, the algorithm
attempts to chooses the function $f_n \in \Fcal$ which it considers as
the best classifier in $\Fcal$, based on the given training points. On the other
hand, in theory we know precisely what the best classifier in $\Fcal$
is: it is the one that has the smallest risk. For simplicity, we assume that it is unique and denote it as $f_{\Fcal}$, that is 
\banum \label{eq-ffcal}
f_{\Fcal} = \argmin_{f \in \Fcal} R(f).
\eanum
The third classifier we will talk about is
the Bayes classifier $f_{Bayes}$ introduced in Equation~\eqref{eq-fbayes} above. This is the best
classifier which exists at all. In the notation above we could also
denote it by $f_{\Fcal_{all}}$ (recall the notation $\Fcal_{all}$ for
the space of all functions). But as it is unknown to the learner,
it might not be contained in the function space $\Fcal$ under
consideration, so it is very possible that $R(f_{\Fcal}) >
R(f_{Bayes})$. With the notation for these three classifiers $f_n$,
$f_{\Fcal}$, and $f_{Bayes}$ we can now define different types of
convergence:

\begin{definition}
Let $(X_i, Y_i)_{i \in \Nat}$ be an infinite sequence of training
points which have been drawn independently from some probability distribution
$P$. Let $\ell$ be a loss function. For each $n \in \Nat$, let $f_n$
be a classifier constructed
by some learning algorithm on the basis of the first $n$ training
points. 

\begin{enumerate}
\item The learning algorithm is called {\em consistent with respect to $\Fcal$
    and $P$}
  if the risk $R(f_n)$ converges in probability to the risk
  $R(f_{\Fcal})$ of the best classifier in $\Fcal$, that is for all
  $\eps > 0$,
\ba
P( R(f_n) - R(f_{\Fcal}) > \eps ) \to 0 \; \text{ as } n \to \infty.
\ea
\item
The learning algorithm is called {\em Bayes-consistent with respect to $P$} if 
 the risk $R(f_n)$ converges to the risk $R(f_{Bayes})$ of the Bayes
classifier, that is for all $\eps > 0$, 
\ba
P( R(f_n) - R(f_{Bayes}) > \eps ) \to 0 \; \text{ as } n \to \infty.
\ea
\item The learning algorithm is called {\em universally consistent with
    respect to $\Fcal$ (resp. universally Bayes-consistent)} if it is
  consistent with respect to $\Fcal$ (resp. Bayes-consistent) for all
  probability distributions $P$.
\end{enumerate}
\end{definition}

Note that for simplicity in the following we often say ``the classifier
$f_n$ is consistent'', meaning that ``the classification algorithm
which, based on $n$ samples, chooses $f_n$ as classifier is consistent''. 
Let us try to rephrase the
meaning of those definitions in words. We start with Part 1 of the
definition.  The statement requests that the larger the sample size
$n$ gets, the closer the risk of the classifier $f_n$ should get to
the risk of the best classifier $f_{\Fcal}$ in the space $\Fcal$.
This should happen ``with high probability'': Note that the risk
$R(f_n)$ is a random quantity, as it depends on the underlying
sample. In rare circumstances (with probability $< \delta$, where
$\delta$ is supposed to be a  small, positive number), it might be
the case that we get mainly misleading sample points (for example,
lots of points with ``wrong'' labels due to label noise). In those
circumstances, it might happen that our classifier is not very
good. However, in the vast majority of all cases (with probability $\ge
1-\delta$), our training points will not be misleading, at least if we
have many of them ( $n \to \infty$). Then the classifier $f_n$ picked
by the algorithm will be close to the best classifier the
algorithm could have picked at all, the classifier
$f_{\Fcal}$, that is $R(f_n) - R(f_{\Fcal}) > \eps$ only with small
probability $\delta$, where $\delta$ will converge to 0 as $n \to \infty$.\\

Part 2 of the definition is similar to Part 1, except that we now
compare to the overall best classifier $f_{Bayes}$. The difference
between those statements is obvious: Part 1 deals with the best the
algorithm can do under the given circumstances (namely, in the
function space $\Fcal$). Part 2 compares this to the overall best
possible result. Traditionally, statistical learning theory often
focuses on Part 1, but ultimately we will be more interested in Part
2. Both parts will be treated in the following sections.  \\

To understand Part 3, note the following. In the first two parts,
consistency is defined for a fixed probability distribution $P$. This
means that if the true underlying distribution is $P$, then the
sequence of classifiers $f_n$  will converge to the correct result. However, the whole
point of machine learning is that we do not know what the
underlying distribution $P$ is. So it would make little sense if
a learning algorithm is consistent for a certain distribution $P$, but
inconsistent for some other distribution $P'$. Hence, we define a
stronger notion of consistency, termed universal consistency. It states
that no matter what the underlying distribution might be, we will
always have consistency. \\

A mathematical detail we would like to skim over is the exact type of
convergence. For readers familiar with probability theory: consistency
as stated above is called weak consistency, as it is a statement about
convergence in probability; the analogous statement for convergence
almost surely would be called
strong consistency. For exact details see Section~6  in
\citet{DevGyoLug96}. \\

There is one further important fact to note about these definitions:
they never mention  the empirical risk $\remp(f_n)$ of a classifier,
but are only concerned with the true risk $R(f_n)$. On the one hand,
it is clear why this is the case: our measure of quality of a
classifier is the true risk, and we want the true risk to become as
good as possible. On the other hand, the empirical risk is our first
and most important estimator of the true risk of a classifier. So it
seems natural that in addition to the convergence of the true risk such
as $R(f_n) \to R(f_{Bayes})$, we also request convergence of the
empirical risk: $\remp(f_n) \to R(f_{Bayes})$. We will see below that
such statements about the empirical risk are the most important steps
to prove consistency in the standard approach to statistical learning
theory.  So even though we did not explicitly require convergence of
the empirical risk, it usually comes out as a side result of consistency.

\subsection{The bias-variance and estimation-approximation trade-off} \label{subsec-bias-variance}

\begin{figure}[t]
  \begin{center}
 \includegraphics[width=0.4\textwidth]{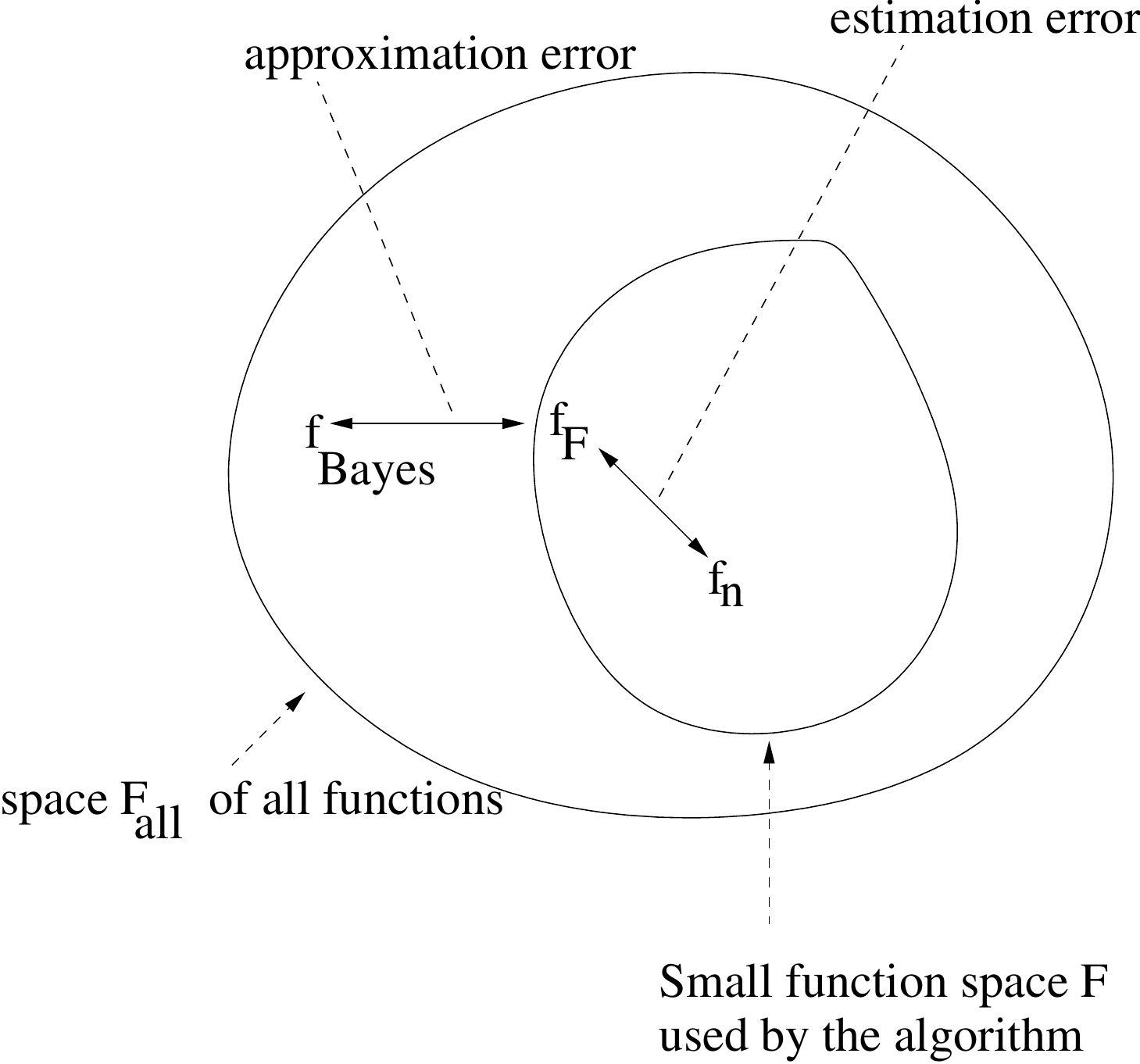}
\caption{\em Illustration of estimation and approximation error.}
  \label{fig-est-approx}
  \end{center}
\end{figure}

The example illustrated in Figure~\ref{fig:slt:fig1} above already
pointed out the problem of model complexity in an intuitive way: when
is a model ``simpler'' than another one? Is it good that a model is
simple? How simple? %
We have already stated above that the goal of classification
is to get a risk as good as the Bayes classifier. Could we just
choose $\Fcal$ as the space $\Fcal_{all}$ of all functions, define the
classifier $f_n := \argmin_{f \in \Fcal_{all}} \remp(f)$, and obtain
consistency? Unfortunately, the answer is no. In the sections below we
will see that if we optimize over too large function classes $\Fcal$,
and in particular if we make $\Fcal$ so large that it contains all the
Bayes classifiers for all different probability distributions $P$,
this will lead to inconsistency. So if we want to learn successfully,
 we need to work with a smaller function class
$\Fcal$. To investigate the competing properties of model complexity
and generalization, we want to introduce a few notions which will be
helpful later on. \\

Recall the definitions $f_n$, $f_{\Fcal}$ and $f_{Bayes}$ introduced
above. We have seen that Bayes-consistency deals with the convergence of the
term $R(f_n) - R(f_{Bayes})$. Note that we can decompose this quantity in the
following way:

\banum \label{eq-est-approx}
R(f_n) - R(f_{Bayes}) = 
\underbrace{\Big(R(f_n) -  R(f_{\Fcal})\Big)}_{\text{estimation error}} \; + \;   
\underbrace{\Big( R(f_{\Fcal}) - R(f_{Bayes})
  \Big)}_{\text{approximation error}}
\eanum

The two terms on the right hand side have particular names: the first
one is called the {\em estimation error} and the second one the {\em
  approximation error}; see also Figure \ref{fig-est-approx}
for an
illustration. 
The reasons for these names are as follows. The first term deals with
the uncertainty introduced by the random sampling
process. That is, given the finite sample, we need to estimate
the best function in $\Fcal$. Of course, in this process we will make some
(hopefully small) error. This error is called the {\em estimation error}. 
The second term is not influenced by any random quantities. It deals
with the error we make by looking for the best function in a (small)
function space $\Fcal$, rather than looking for the best function in
the entire space $\Fcal_{all}$  of all functions. The fundamental question in this context
is how well  functions in $\Fcal$ can be used to approximate
functions  $\Fcal_{all}$ in the space of all functions. Hence the name
{\em approximation error}. \\

In statistics,
estimation error is also called the {\em variance}, and the approximation
error is called the {\em bias} of an estimator. Originally, these
terms were coined for the special situation of regression with squared error
loss, but by now people use them in more general settings, like  the
one outlined above. The intuitive meaning is the same: the first term
measures the variation of the risk of the function $f_n$ estimated on
the sample, the second one measures the ``bias'' introduced in the
model by choosing 
too small a function class. \\

\begin{figure}[t]
  \begin{center}
 \includegraphics[width=0.4\textwidth]{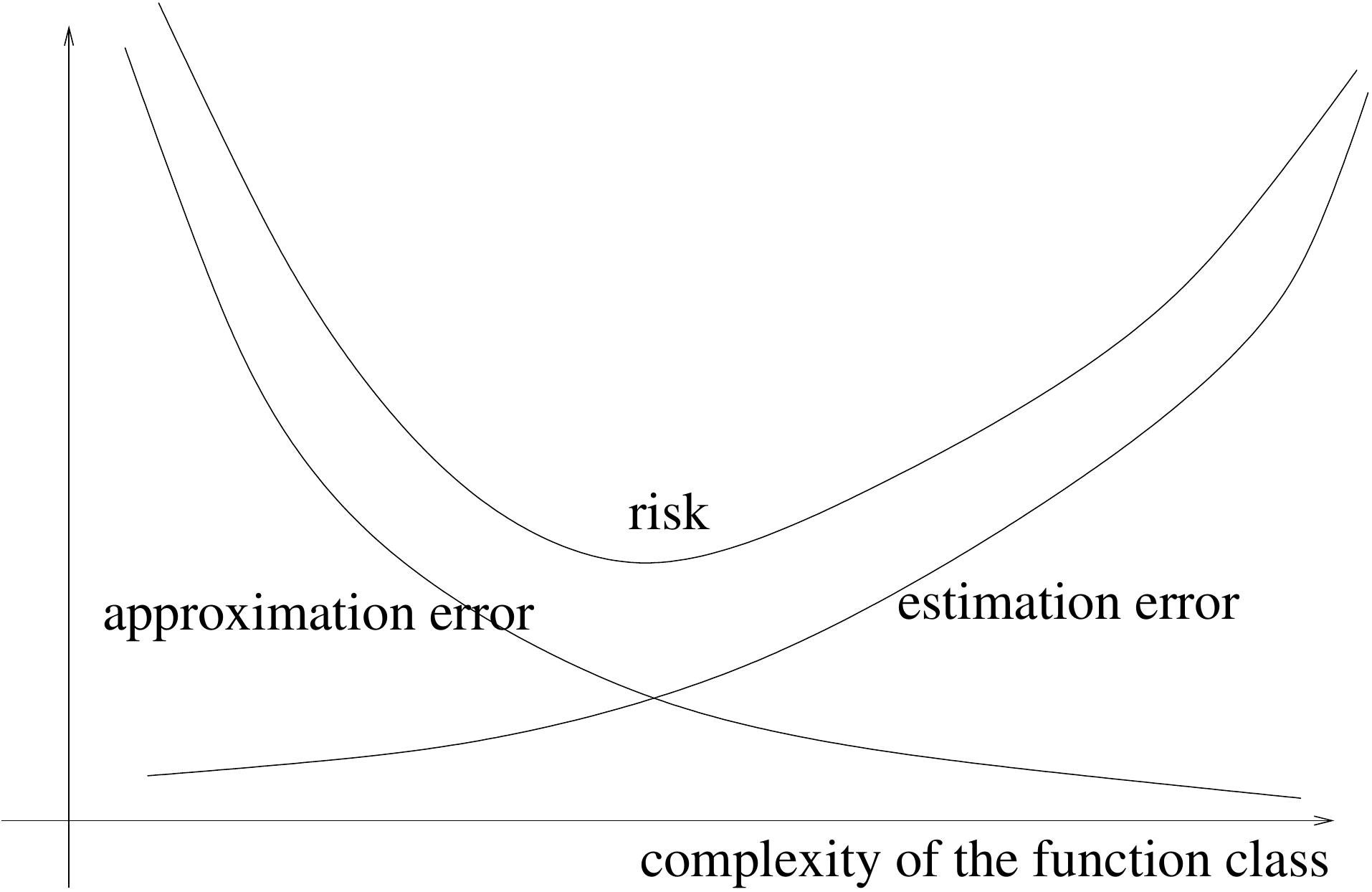}
\caption{\em The trade-off between estimation and approximation error. If the
  function space $\Fcal$ used by the algorithm has a small complexity,
then the estimation error is small, but the approximation error is
large (underfitting). If the complexity of $\Fcal$ is large, then the
estimation error is large, while the approximation error is small
(overfitting). The best overall risk is achieved for  ``moderate''
complexity. }
  \label{fig-est-approx2}
  \end{center}
\end{figure}

At this point, we would already like to point out that the space
$\Fcal$ is the means to balance the trade-off between estimation and approximation
error; see Figure~\ref{fig-est-approx2} for an illustration and
Sections~\ref{sec-erm} and \ref{sec-capacity} for an in-depth
discussion. If we choose a very large space
$\Fcal$, then the approximation term will become small (the
Bayes classifier might even be contained in $\Fcal$ or can be
approximated closely by some element in $\Fcal$). The estimation
error, however, will be rather large in this case: the space $\Fcal$
will contain complex functions which will lead to
overfitting. The opposite effect will happen if the function class
$\Fcal$ is very small.\\

In the following, we will deal with the estimation error and
approximation error separately. We will see that they have rather
different behavior and that different methods are needed to control
both. Traditionally, SLT has a strong focus on the estimation
error, which we will discuss in greater depth in Sections~\ref{sec-erm} and
\ref{sec-capacity}. The
approximation error will be treated in Section~\ref{sec-approx-error}.

\section{Consistency and generalization for the \emph{k}-nearest neighbor
  classifier} \label{sec-knn}

For quite some time, until 1977, it was not known whether a
universally consistent classifier exists. This question has been
solved positively by \citet{Stone77} who showed by an elegant proof
that a particular classifier, the so-called $k$-nearest neighbor
classifier, is universally consistent. As the $k$-nearest neighbor
classifier is one of the simplest classifiers and is
still widely used in practice, we would
like to spend this section illustrating the notions introduced in
the last section such as
generalization, overfitting, underfitting, and consistency at the
example of the $k$-nearest neighbor classifier. \\

Assume we are given a sample of points and labels $(X_1, Y_1), ...,
(X_n, Y_n)$ which live in some metric space. This means that we have
some way of computing distances between points in this space. Very
generally, the paradigm of learning is to assign ``similar output to
similar inputs''. That is, we believe that points which are ``close''
in the input space tend to have the same label in the output
space. Note that if such a statement does not hold, learning becomes
very difficult or even impossible. For successful learning, there
needs to be some way to relate the labels of training points to those
of test points, and this always involves some prior assumptions about
relations between input points. The easiest such relation is a
distance between points, but other ways of measuring similarity,
such as ``kernels,'' exist and indeed form the basis of some of 
the most popular existing learning algorithms \citep{SchSmo02}. \\

So assume that there exists a distance function on the input space,
that is a function $d:\Xcal \times \Xcal \to \R$ which assigns a
distance value $d(X, X')$ to each pair of training points $X,
X'$. Given some training points, we now want to predict a good label for a new test point
$X$. A simple idea is  to search for the training point $X_i$ which has the smallest
distance to $X$, and then give $X$ the corresponding label $Y_i$ of
that point. To define this more formally, denote by $\nn(X)$ the
nearest neighbor of $X$ among all training points, that is 

\ba
\nn(X) = \argmin\{X' \in \{X_1, ..., X_n\} \condon d(X, X') \leq d(X,
  X'') \text{ for all } X'' \in  \{X_1, ..., X_n\} \}. 
\ea

We can then define the classifier $f_n$ based on the
sample of $n$ points by 

\ba
f_n(X) = Y_i \text{ where } X_i = \nn(X).
\ea

This classifier is also called the 1-nearest neighbor
classifier (1NN classifier). 
A slightly more general version is the $k$-nearest neighbor
classifier (kNN classifier). Instead of just looking at the closest training point, we
consider the closest $k$ training points, and then take the average
over all their labels. That is, we define the $k$-nearest neighbors
$\knn(X)$ of
$X$ as the set of those $k$ training points which are closest to
$X$. Then we set the kNN classifier 

\ba
f_n(X) = 
\begin{cases}
+1 & \text{ if } \sum_{X_i \in \knn(X)} Y_i > 0\\
-1 & \text{ otherwise. }
\end{cases}
\ea

That is, we decide on the label of $X$ by  majority vote among the
labels of the training points in the $k$-nearest neighborhood of
$X$. To avoid ties one usually chooses $k$ as an uneven number. \\

Let us first consider the simpler case of the 1-nearest neighbor
classifier. Is this classifier Bayes-consistent? Unfortunately, the
answer is no. To see this, we consider the following counter-example: 

Assume our data space is the real interval $\Xcal = [0,1]$. As
distribution $P$ we choose the distribution which gives uniform weight
to all points $X \in [0,1]$, and which assigns labels with noise such
that $P(Y=1 \condon X = x) = 0.9$ for all $x \in \Xcal$. That is, the
``correct'' label (the one assigned by the Bayes classifier) is +1 for
all points $x \in \Xcal$. We have already mentioned this example when
we introduced the Bayes classifier. In this case, the Bayes classifier
is simply the function which outputs $1$ for all points in $\Xcal$,
and its Bayes risk with respect to the 0-1-loss is 0.1. Now let
us investigate the behavior of the 1NN classifier in this case. When
we draw some training points $(X_i, Y_i)_{i=1,...,n}$ according to the
underlying distribution, then they will be roughly ``evenly spread''
on the interval $[0,1]$. On average, every 10th point will have training
label -1, all others have training label +1. If we now consider the behavior of
the 1NN classifier $f_n$ on this example, we can write its risk with
respect to the 0-1-loss function as follows:

\ba
R(f_n) 
&= 
P( Y \neq f_n(X)) 
= 
P(Y = 1 | f_n(X) = 0 ) + P(Y = 0 | f_n(X) = 1) \\
& \approx
0.1 \cdot 0.9 + 0.9 \cdot 0.1 = 2 \cdot 0.1 \cdot 0.9 = 0.18
\ea

The approximation sign $\approx$ is used because in this argument we
suppress the variations introduced by the random sampling process to
keep things simple. 
We can see that the risk $R(f_n)$ of the classifier $f_n$ is approximately 0.18,
independently of the sample size $n$, while the Bayes
risk would be 0.1. Hence, the 1NN classifier is not consistent as
$R(f_n) \not\to R(f_{Bayes})$.  On the other hand, note that if we
consider the 100-nearest neighbor classifier instead of the 1-nearest
neighbor classifier in this example, we would make much fewer
mistakes: it is just very unlikely to have a neighborhood of 100
points in which the majority vote of training points is -1. Thus, the
100-nearest neighbor classifier, while still not being consistent,
makes a smaller error than the 1-nearest neighbor classifier. \\

The trick to achieve consistency is related to this
observation. Essentially, one has to allow that the size $k$ of the
neighborhood under consideration grows with the sample size
$n$. Formally, one can prove the following theorem: 

\begin{theorem}[\citealp{Stone77}] Let $f_n$ be the $k$-nearest
  neighbor classifier constructed on $n$ sample points. If $n \to
  \infty$ and $k \to \infty$ such that $k/n \to 0$, then
  $R(f_n) \to R(f_{Bayes})$ for all probability distributions $P$. That is,
  the $k$-nearest neighbor classification rule universally Bayes-consistent. 
\end{theorem}

Essentially, this theorem tells us that if we choose the neighborhood
parameter $k$ such that it grows ``slowly'' with $n$, for example $k
\approx \log(n)$, then the $\knn$ classification rule is universally Bayes-consistent. \\

In the last sections we mentioned that the function class $\Fcal$ from
which the classifier is chosen is 
 an important ingredient
for statistical learning theory. In the case of the $\knn$ classifier,
this is not as obvious as it will be for the classifiers we are going
to study in later sections. Intuitively, one can say that for a fixed
parameter $k$, the function
class $\Fcal_k$ is a space of piecewise constant functions. The larger
$k$ is, the larger the $k$-neighborhoods become and thus the larger
the pieces of functions which have to be constant. This means that for
very large $k$, the function class $\Fcal_k$ is rather ``small'' (the
functions cannot ``wiggle'' so much). In the extreme case of $k=n$,
the $k$-neighborhood simply includes all training points, so the $\knn$
classifier cannot change its sign at all --- it has to be constant on the
whole input space $\Xcal$. In this case, the function class $\Fcal_k$
contains only two elements: the function which is constantly +1 and the
function which is constantly -1. On the other hand, if $k$ is small
then $\Fcal_k$ becomes rather large (the functions can change their labels
very often and abruptly). In the terms explained in the last
sections, one can say that if we choose $k$ too small, then the
function class overfits: for example, this happens in the extreme case
of the 1NN classifier. On the other hand, if $k$ is too large, then
the function class underfits --- it simply does not contain functions
which are able to model the training data. \\

In this section we merely touched on the very basics of the $\knn$
classifier and did not elaborate on any proof techniques. For a very
thorough treatment of theoretical aspects of $\knn$ classifiers we
recommend the monograph \citet{DevGyoLug96}.

\section{Empirical risk minimization}\label{sec-erm}

In the previous section we encountered our first simple classifier: the
$\knn$ classifier. In this section we now want to turn to a more
powerful way of classifying data, the so called empirical risk
minimization principle. 
Recall the assumption that
the data are generated iid (independent and identically distributed)
from an unknown underlying distribution $\Pr(X,Y)$. As we have already
seen, the learning
problem consists in minimizing the \emph{risk} (or \emph{expected
  loss} on the test data),
\begin{equation}
\label{eq:slt:r}
R(f)= E ( \ell(X, Y, f(X))
\end{equation}
where $f$ is a function mapping the input space $\Xcal$ into the label
space $\Ycal$, and $\ell$ is the
loss function. 
The difficulty of this task stems from the fact that we are trying to
minimize a quantity that we cannot actually evaluate: since we do not
know the underlying probability distribution $\Pr$, we cannot compute the risk $R(f)$. What we
\emph{do} know, however, are the training data %
sampled from $\Pr$.  We can thus try to infer a function $f$ from the
training sample whose risk is \emph{close} to the best possible risk. To this
end, we need what is called an \emph{induction
  principle}.\index{induction~principle} \\

Maybe the most straightforward way to proceed is to approximate the
true risk by the empirical risk computed on the training data. Instead
of looking for a function which minimizes the true risk $R(f)$, we
try to find the one which minimizes the empirical risk 
\begin{equation}
\label{eq:slt:r_emp}
R_\emp(f)=\frac{1}{n}\sum_{i=1}^n \ell(X_i, Y_i, f(X_i)).
\end{equation}
That is, given some training data $(X_1, Y_1), ..., (X_n, Y_n)$, a
function space $\Fcal$ to work with, and a loss function $\ell$, 
we define the classifier $f_n$ as the function 
\ba
f_n := \argmin_{f \in \Fcal} \Remp(f).
\ea
This approach is called the \emph{empirical risk minimization induction
principle}, abbreviated by ERM. The motivation for this principle is given by
the law of large numbers, as we will now explain. \\

\subsection{The law of large numbers}
\label{sec:slt:largenumbers}

The law of large numbers is one of the most important theorems in
statistics. In its simplest form it states that under mild conditions,
the mean of random variables $\xi_i$ which have been
drawn iid from some probability distribution $P$ converges to the mean
of the underlying distribution itself when the sample size goes to
infinity: 
\ba
\frac{1}{n} \sum_{i=1}^n \xi_i \to E ( \xi) \;\; \text{ for } n \to
\infty.
\ea
Here the notation assumes that the sequence $\xi_1, \xi_2, ...$ has been
sampled iid from $P$ and that $\xi$ is also distributed according to
$P$.  This theorem can be applied to the case of the empirical and
the true risk. In order to see this, note that the empirical risk is
defined as the mean of the loss $\ell(X_i,Y_i,f(X_i))$ on individual
sample points, and the true risk is the mean of this loss over the
whole distribution. That is, from the law of large numbers we can
conclude that for a fixed function $f$, the empirical risk converges
to the true risk as the sample size goes to infinity: 
\ba
\remp(f) = \frac{1}{n} \sum_{i=1}^n \ell(X_i, Y_i, f(X_i)) \to E ( \ell(X,Y,f(X))) \;\; \text{ for } n \to
\infty.
\ea
Here the loss function $\ell(X,Y,f(X))$ plays the role of the random
variable $\xi$ above. For a given,
finite sample this means that we can approximate the true
risk  (the one we are interested in) very well by the
empirical risk (the one we can compute on the
sample).
A famous inequality due to %
\citet{Chernoff52}, later generalized by \citet{Hoeffding63},
characterizes how well the empirical mean approximates the expected
value. Namely, if $\xi_i$ are random variables which only take values in the interval $[0,1]$, then
\begin{equation}
  \label{slt:chernoff}
  \Pr \left( \left| \frac{1}{n} \sum_{i=1}^n \xi_i - \Eb(\xi) \right| \ge
    \epsilon \right) \le 2 \exp(-2n\epsilon^2).
\end{equation}
This theorem states that the probability that the sample mean
deviates by more than $\eps$ from the expected value of the distribution
is bounded by a very small quantity, namely by $2
\exp(-2n\epsilon^2)$. Note that the higher $n$ is, the smaller this
quantity becomes, that is the probability for large deviations
decreases very fast with $n$. Again, we can apply this theorem to the
setting of empirical and true risk. This leads to 
a
bound which states how likely it is that for a given function $f$ the
empirical risk is close to the actual risk: 
\begin{equation}\label{slt-eq:lawoflage}
\Pr ( |R_\emp(f) - R(f)| \ge \epsilon ) \le 2\exp(-2n\epsilon^2).
\end{equation}
For any fixed function (and sufficiently large \textit{n}), it is thus
highly probable that the training error provides a good estimate of
the test error. \\

There are a few  important facts concerning the Chernoff bound 
\eq{slt-eq:lawoflage}. First, a crucial property of the Chernoff bound
is that it is probabilistic in nature. It states that the probability
of a large deviation between the test error and the training error of $f$ is
small; the larger the sample size $n$, the smaller the probability.
Hence, it does not rule out the presence of cases where the deviation
is large, it just says that for a fixed function $f$, this is very
unlikely to happen.  The reason why this has to be the case
is the random generation of training points. It could be that
in a few unlucky cases, our training data is so misleading that it is
impossible to construct a good classifier based on it. However, as the
sample size gets larger, such unlucky cases become very rare. In this
sense, any consistency guarantee can only be of the form
``the empirical risk is close to the true risk, with high
probability''. \\

At first sight, it seems that the Chernoff bound
\eq{slt-eq:lawoflage} is enough to prove consistency of empirical risk
minimization.  However, there is an important caveat: the Chernoff bound only
holds for a fixed function $f$ which {\em does not depend on the
  training data.} However, the classifier $f_n$ of course does depend
on the training data (we used the training data to select $f_n$).
While this seems like a subtle mathematical difference, this is where
empirical risk minimization can go completely wrong. We will now discuss
this problem in detail, and will then discuss how to
adapt the strong law of large numbers to be able to deal with
data-dependent functions. 
\subsection{Why empirical risk minimization can be inconsistent} \label{subsec-example-inconsistent}

Assume our underlying data space is $\Xcal = [0,1]$. Choose the  uniform distribution on
$\Xcal$ as probability distribution and define the label $Y$ for an input point $X$ deterministically as follows: 
\begin{equation}
Y = 
\begin{cases}
-1 & \mbox{~if~} X < 0.5 \\
1   & \mbox{~if~} X \geq 0.5.
\end{cases}
\end{equation}
Now assume we are given a set of training points $(X_i,
Y_i)_{i=1,...,n}$, and consider the
following classifier: 
\begin{equation}
f_n(X) = \left\{
\begin{array}{ll}
Y_i & \mbox{~if~} X = X_i \mbox{~for some~} i=1,\dots,n\\
1   & \mbox{~otherwise.}
\end{array}
\right.
\label{eq:example}
\end{equation}
This classifier $f_n$ perfectly classifies all training points. That
is it has empirical risk $\remp(f_n) =0$. Consequently, as the
empirical risk cannot become negative, $f_n$ is a minimizer of the
empirical risk.  However, $f_n$ clearly has not learned anything, the
classifier just memorizes the training labels and otherwise simply
predicts the label $1$. Formally, this means that the classifier $f_n$
will not be consistent. To see this, suppose we are given a test point
$(X,Y)$ drawn from the underlying distribution.  Usually, this test
point will not be identical to any of the training points, and in this
case the classifier simply predicts the label $1$. If $X$ happened to
be larger than $0.5$, this is the correct label, but if $X < 0.5$, it
is the wrong label. Thus the classifier $f_n$ will err on half of all
test points, that is it has test error $R(f_n) = 1/2$. This is the
same error one would make by random guessing!  In fact, this is a nice
example of overfitting: the classifier $f_n$ fits perfectly to the
training data, but does not learn anything about new test data. It is
easy to see that the classifier $f_n$ is inconsistent. Just note that
as the labels are a deterministic function of the input points, the
Bayes classifier has risk 0. Hence
we have $1/2 = R(f_n) \not\to R(f_{Bayes}) = 0$. \\

We have constructed an example where empirical risk minimization
fails miserably. Is there any way we can rescue the ERM principle?
Luckily, the answer is yes. The main object we have to look at is the
function class $\Fcal$ from which we draw our classifier. %
If we allow our
function class to contain functions which just memorize  the
training data, then the ERM principle cannot work. In particular,  if we  choose the empirical
risk minimizer from  the space
$\Fcal_{all}$ of all functions between $\Xcal$ and $\Ycal$, then the values of the function $f_n$ at the training points 
points $X_1,\dots,X_n$  do not necessarily carry any information about the values at other
points.
Thus, unless we make restrictions on the space of functions from which
we choose our estimate $f$, we cannot hope to learn anything.
Consequently, machine learning research has studied various ways to
implement such restrictions. In statistical learning theory, these
restrictions are enforced by taking into account the \emph{capacity}
of the space of functions that the learning machine can implement.
\subsection{Uniform convergence} \label{subsec-uniform}

It turns out the conditions required to render empirical risk
minimization consistent involve \emph{restricting the set of
  admissible functions}.  The main insight of VC (Vapnik-Chervonenkis)
theory is that the consistency of
empirical risk minimization is determined by 
 the \emph{worst case} behavior over all functions
$f \in \Fcal$  that
the learning machine could choose. We will see that instead of the standard
law of large numbers introduced above, this worst case corresponds to
a version of the
law of large numbers which is {\em uniform} over all functions in
$\Fcal$. 
\begin{figure}[t]
  \centerline{\includegraphics[width=0.6\textwidth]{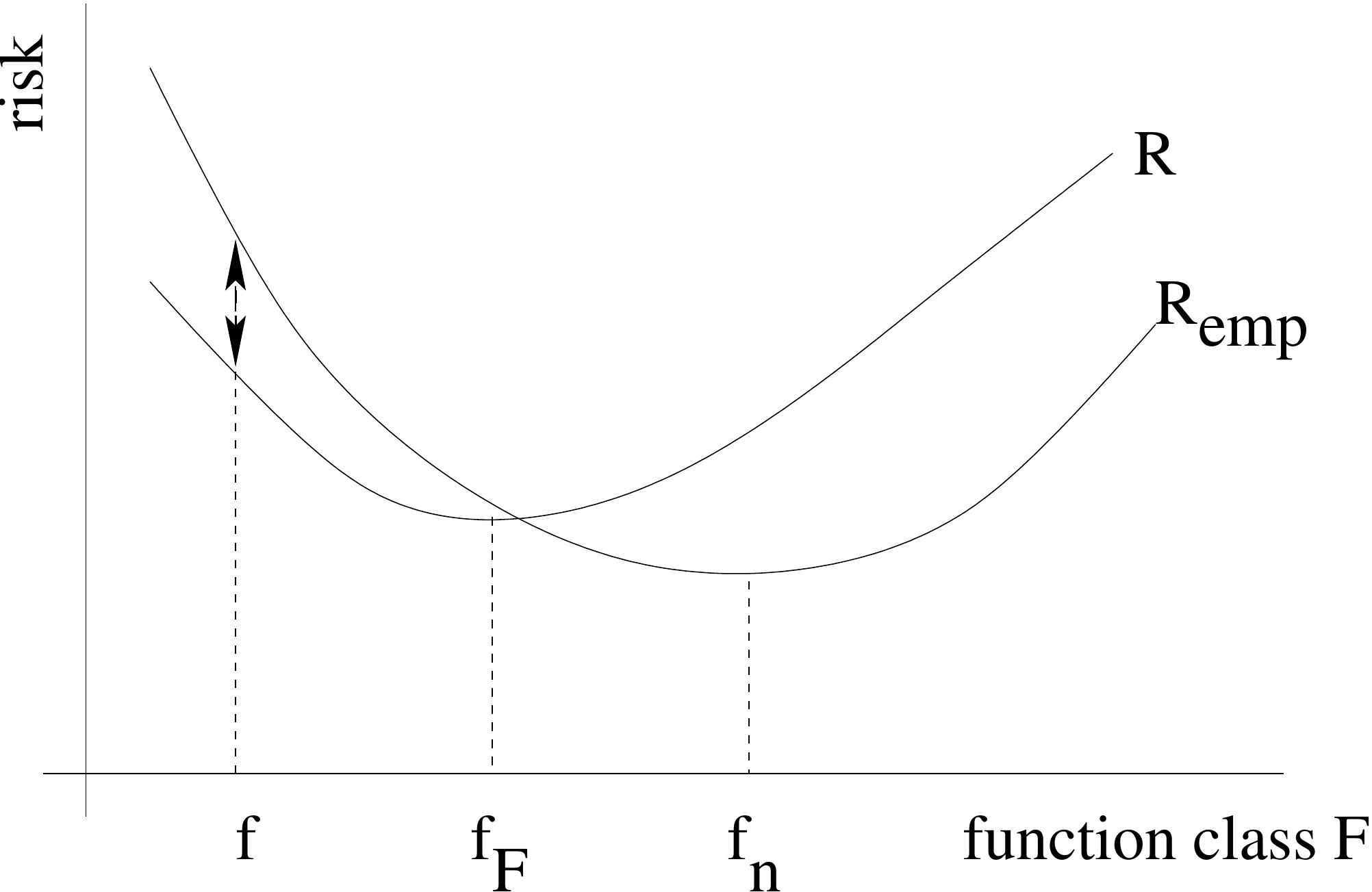}}
\caption{ \em 
Simplified depiction of the convergence of empirical risk
to actual risk. The $x$-axis gives a one-dimensional representation of
the function class $\Fcal$ (denoted F in the figure); the $y$ axis denotes the risk. For each
\emph{fixed} function $f$, the law of large numbers tells us that as
the sample size goes to infinity, the empirical risk
$R_{\emp}(f)$ converges towards the true risk $R(f)$
(indicated by the arrow). This does not imply, however, that
in the limit of infinite sample sizes, the {\em minimizer} of the empirical
risk, $f_n$, will lead to a value of the risk that is as good as the
risk of the best function $f_{\Fcal}$ in
the function class  (denoted $f_F$ in the figure). For
the latter to be true, we require the convergence of
$R_{\emp}(f)$ towards $R(f)$ to be uniform over all functions in
$\Fcal$ (from \citealp{SchSmo02}).
\label{slt-fig:uniform}
}
\end{figure}
\fig{slt-fig:uniform} gives a simplified depiction of the uniform law
of large numbers and the question of consistency.  Both the empirical
risk and the actual risk are plotted as functions of $f$. For
simplicity, we have summarized all possible functions $f$ by a single
axis of the plot.  Empirical risk minimization consists in picking the
$f$ that yields the minimal value of $R_\emp$. It is consistent if the
minimum of $R_\emp$ converges to that of $R$ as the sample size
increases. One way to ensure the convergence of the minimum of all
functions in $\Fcal$ is uniform convergence over $\Fcal$: we require
that for {\em all} functions $f \in \Fcal$, the difference between
$R(f)$ and $\Remp(f)$ should become small {\em simultaneously}. That
is, we require that there exists some large $n$ such that for sample
size at least $n$, we know {\em for all functions $f \in \Fcal$} that
$|R(f) - \remp(f)|$ is smaller than a given value $\eps$. In the
figure, this means that the two plots of $R$ and $\remp$ become so
close that their distance is never larger than $\eps$.
Mathematically, the statement ``$|R(f) - \remp(f)| \leq \eps$ for all
$f \in\Fcal$'' can be expressed using a supremum (readers not familiar
with the notion of ``supremum'' should think of a maximum instead):
\ba
\sup_{f \in \Fcal} |R(f) - \remp(f)| \leq \eps. 
\ea
Intuitively it is clear that if we know that for all functions $f \in
\Fcal$ the difference $|R(f) - \remp(f)|$ is small, then this holds in
particular for any function $f_n$ we might have chosen based on the
given training data. That is, for any function $f \in \Fcal$ we have

\ba
|R(f) - \Remp(f)| \;\;\leq\;\; \sup_{f \in \Fcal} |R(f) - \remp(f)|. 
\ea
In particular, this also holds for a function $f_n$ which has been
chosen based on a finite sample of training points. 
From this we can draw the following conclusion: 

\banum \label{eq-uniform}
P( |R(f_n) - \Remp(f_n) | \geq \eps ) \;\; \leq \;\; 
P( \sup_{f \in \Fcal} |R(f) - \remp(f)| \geq \eps ).
\eanum

The quantity on the right hand side is now what the uniform law of
large numbers deals with.  We say that the
law of large numbers holds uniformly over a function class $\Fcal$ if
for all $\eps > 0$, 

\ba
P( \sup_{f \in \Fcal} |R(f) - \remp(f)| \geq \eps ) \to 0 \text{ as }
n \to \infty.
\ea

One can now use \eqref{eq-uniform} to show that if the uniform law of
large numbers holds for some function class $\Fcal$, then empirical
risk minimization is consistent with respect to $\Fcal$. 
To see this, consider the following derivation: \\

\ba
& |R(f_n) - R(f_\Fcal)| \\
& \note{by definition of $f_\Fcal$ we know that $R(f_n) - R(f_\Fcal)
  \geq 0$}\\
& = 
R(f_n) - R(f_\Fcal)\\
& = 
R(f_n) - \remp(f_n) 
+ \remp(f_n) - \remp(f_\Fcal) 
+ \remp(f_\Fcal) - R(f_\Fcal)\\
& \note{note that $\remp(f_n) - \remp(f_\Fcal) \leq 0$  by
    def. of  $f_n$}\\
& \leq 
R(f_n) - \remp(f_n) + \remp(f_\Fcal) - R(f_\Fcal)\\
& \leq 2 \sup_{f \in \Fcal} | R(f) - \remp(f)|
\ea

So we can conclude: 

\ba 
P( |R(f_n) - R(f_\Fcal)| \geq \eps) 
\;\;\leq\;\; 
P( \sup _{f \in \Fcal} | R(f) - \remp(f)| \geq \eps/2). 
\ea

Under the uniform law of large numbers, the right hand side
tends to 0, which then leads to consistency of ERM with respect to the
underlying function class $\Fcal$. 
In other words, uniform convergence over $\Fcal$ is a sufficient condition for
consistency of the empirical risk minimization over $\Fcal$. \\

What about the other way round? Is uniform convergence also
a \emph{necessary} condition? Part of the elegance of VC
theory lies in the fact that this is the case (see for example 
\citealp{VapChe71}, \citealp{Mendelson03}, \citealp{DevGyoLug96}):

\begin{theorem}[Vapnik \& Chervonenkis]\label{slt-theo:unifo}
Uniform convergence 
\begin{equation}
\Pr(\sup_{f\in\Fcal} |R(f) - R_\emp(f)|>\epsilon) \to 0 \text{ as }
n \to \infty,
\end{equation}
for all $\epsilon>0$, is a \emph{necessary and sufficient} condition for
consistency of empirical risk minimization with respect
to $\Fcal$.
\end{theorem}

In Section \ref{subsec-example-inconsistent} we gave an example
where we considered the set of all possible functions, and showed that
learning was impossible.  The dependence of learning on the underlying
set of
functions has now returned in a different guise: the condition of
uniform convergence crucially depends on the set of functions for
which it must hold. Intuitively, it seems clear that the larger the
function space $\Fcal$, the larger 
$\sup_{f\in\Fcal} |R(f) - R_\emp(f)|$. Thus, the larger $\Fcal$, the larger  
$\Pr ( \sup_{f\in\Fcal} |R(f) - R_\emp(f)|>\epsilon)$. Thus, the
larger $\Fcal$, the more difficult it is to satisfy the uniform law of
large numbers. That is, for larger function spaces $\Fcal$ consistency
is ``harder'' to achieve than for ``smaller'' function spaces. \\

This abstract characterization of consistency as a uniform convergence
property, whilst theoretically intriguing, is not all that useful in
practice. The reason is that it seems hard to find out whether the
uniform law of large numbers holds for a particular function class
$\Fcal$. Therefore, we next address whether there are properties of
function spaces which \emph{ensure} uniform convergence of risks.

\section{Capacity concepts and generalization bounds} \label{sec-capacity}

So far, we have argued which property of the function space determines
whether the principle of empirical risk minimization is consistent,
i.e., whether it will work ``in the limit.'' This property was
referred to as uniform convergence. To make statements about what
happens after seeing only finitely many data points --- which in
reality will always be the case --- we need to take a closer look at
this convergence. It will turn out that this will provide us with
bounds on the risk, and it will also provide insight into which
properties of function classes determine whether uniform convergence
can take place. To this end, let us take a closer look at the
subject of Theorem~\ref{slt-theo:unifo}: the probability
\begin{equation}\label{slt-eq:the-event}
\Pr(\sup_{f\in\Fcal} | R(f)-R_\emp(f)| >\epsilon).
\end{equation}

Two tricks are needed along the way: the \emph{union bound} and the
method of \emph{symmetrization by a ghost sample}.

\subsection{The union bound} \label{subsec-union-bound}
The union bound is a simple but convenient tool to transform the
standard law of large numbers of individual functions into a uniform
law of large numbers over a set of finitely many functions. 
Suppose the set $\Fcal$ consists just of finitely many functions,
that is $\Fcal = \{f_1, f_2, ..., f_m\}$. Each of the functions $f_i
\in \Fcal$ satisfies the standard  law
of large numbers in form of the Chernoff bound, that is 
\banum \label{eq-chernoff}
& P( |R(f_i) - \remp(f_i)| \geq \eps) \leq 2\exp(-2n \eps^2). \\
\eanum 
Now we want to transform these statements about the individual
functions $f_i$ into a uniform law of large numbers. To this end, note
that we can rewrite: 
\banum
& \Pr(\sup_{f\in\Fcal} |R(f)-R_\emp(f)| \geq \eps ) \nonumber \\
& = 
\Pr\Big( \;
|R(f_1) -R_\emp(f_1)| \geq \eps  \; \text{ or }\; 
|R(f_2) -R_\emp(f_2)| \geq \eps    \; \text{ or }\;  ...  \; \text{ or }\; 
|R(f_m) -R_\emp(f_m)| \geq \eps   \;\Big) \nonumber \\
& 
\leq \sum_{i=1}^m \Pr(|R(f_i) -R_\emp(f_i)| \geq \eps ) \nonumber\\
& 
\leq 2m \exp(-2n \eps^2) \label{eq-union-bound}
\eanum
Let us go through these calculations step by step. The first equality
comes from the way the supremum is defined. Namely, the supremum over
certain expressions is larger than $\eps$ if at least one of the
expressions is larger than $\eps$, which leads to the statements with
the ``or'' combinations. The next step uses a standard tool from probability theory, the union
bound. The union bound states that the probability of a union of
events (that is, events coupled with ``or'') is smaller or
equal to the sum of the individual probabilities. Finally, the last
step is a simple application of the Chernoff bound of  Eq. \eqref{eq-chernoff} to each of the
terms in the sum. \\

From left to right, the statements in Eq. \eqref{eq-union-bound} show
us how to convert the Chernoff bound for individual functions $f_i$
into a bound which is uniform over a finite number $m$ of
functions. As we can see, the difference between the Chernoff bound
for the individual functions and the right hand side of
\eqref{eq-union-bound} is just a factor $m$. If the function space
$\Fcal$ is fixed, this factor can be regarded as a  constant, and the term $2m \exp(-2n \eps^2)$ still
converges to 0 as $n \to \infty$. Hence, the empirical risk converges
to 0 
uniformly over $\Fcal$ as $n \to \infty$.  That is, we have
proved that empirical risk minimization over a {\em finite } set $\Fcal$ of
functions is consistent with respect to
$\Fcal$. \\

We next describe a trick used by Vapnik and Chervonenkis to reduce the
case of an infinite function class to the case of a finite one.  It consists of introducing what is
sometimes called a {ghost sample}. It will enable us to replace
the factor \emph{m} in \eq{eq-union-bound} by more general \emph{capacity measures}
that can be computed for infinite function classes.

\subsection{Symmetrization} \label{subsec-symmetrization}
Symmetrization is an important technical step towards using capacity measures of
function classes. Its main purpose is to replace the event 
$\sup_{f \in \Fcal} | R(f) - \remp(f)|$ by an alternative event which
can be solely computed on a given sample. Assume we are given a sample
$(X_i, Y_i)_{i=1,...,n}$.%
We now  introduce a new sample called the {\em ghost
sample}. This ghost sample is just another sample $(X_i',
Y_i')_{i=1,...,n}$ which is also drawn iid from the underlying
distribution and which is independent of the first sample. It is
called ghost sample because we do not need to physically draw this
sample in practice. It is just a mathematical tool, that is we play
``as if we had a second sample''. Of course, given the ghost sample we
can also compute the empirical risk of a function with respect to this
ghost sample, and we will denote this risk by $\remp'(f)$. With the
help of the ghost sample, one can now prove the following simple
statement: 

\begin{lemma}[Vapnik and Chervonenkis]
\label{slt-theo:basiclemma}
  For $m\epsilon^2\ge 2$, we have
  \begin{equation}
    \label{slt-eq:symmetrization}
    \Pr(\sup_{f\in\Fcal} |R(f)-R_\emp(f)|>\epsilon)
\;\; \le \;\;  2 \Pr(\sup_{f\in\Fcal} |R_\emp(f)-R_\emp'(f)|>\epsilon/2).
  \end{equation}
  Here, the first $\Pr$ refers to the distribution of an iid sample of
  size $n$, while the second one refers to the distribution of {\em
    two} samples of size $n$ (the original sample and the ghost
  sample), that is the distribution of iid samples of size $2n$. In
  the latter case, $R_\emp$ measures the empirical loss on the first half of the
  sample, and $R_\emp'$ on the second half.
\end{lemma}

Although we do not prove this lemma, it should be fairly plausible: if
the empirical risks on two independent $n$-samples are close to each
other, then they should also be close to the true risk. \\

This lemma is called the symmetrization lemma. Its name refers to the
fact that we now look at an event which depends in a symmetric way on
a sample, now of size $2n$. 
The main purpose of this lemma is that the quantity $R(f)$, which
cannot be computed on a finite sample, has been replaced by the
quantity $\remp'(f)$, which can be computed on a finite sample. \\

Now let us explain what the symmetrization lemma is used for.  In the
last section we have seen how to bound the probability of uniform
convergence \eq{slt-eq:the-event} in terms of a probability of an
event referring to a \emph{finite} function class. The crucial
observation is now that even if $\Fcal$ contains infinitely many
functions, the different ways it can classify a training set of $n$
sample points is finite. Namely, for any given training point in the
training sample, a function can only take the values $-1$ or $+1$. On
a sample of $n$ points $X_1,\dots,X_n$, a function can act in at most
$2^n$ different ways: it can choose each $Y_i$ as $-1$ or $+1$. This
has a very important consequence. Even if a function class $\Fcal$
contains infinitely many functions, there are at most $2^n$ different
ways those functions can classify the points of a finite sample of $n$
points. This means that if we consider the term 
\ba
\sup_{f\in\Fcal} |R_\emp(f)-R_\emp'(f)| 
\ea
then the supremum effectively only runs over a finite function
class. To understand this, note that two functions $f, g \in \Fcal$
which take the same values on the given sample have the same empirical
risk, that is $\Remp(f) = \Remp(g)$. The analogous statement holds for
the ghost sample and $\Remp'$. Hence, all functions $f, g$ which
coincide both on the sample and the ghost sample will lead to the same
term $|R_\emp(f)-R_\emp'(f)|$. Thus, the only functions we need to
consider to compute the supremum are the $2^{2n}$ functions we can
obtain on sample and ghost sample together. Hence, we can replace the
supremum over $f \in \Fcal$ by the supremum over a finite function
class with at most  $2^{2n}$ functions.\\

Note that this step is only possible due to the symmetrization
lemma. If we had considered the term $\sup_{f \in \Fcal} |R(f) -
\remp(f)|$, the argument from above would not hold, as the value of
$R(f)$ not only depends on the values of $f$ on the sample. \\

In the following we now want to show how the insights gained in the
symmetrization step can be used to derive a first capacity measure of
a function class.

\subsection{The shattering coefficient} \label{subsec-shattering}
For the purpose of bounding \eq{slt-eq:the-event},
Lemma~\ref{slt-theo:basiclemma} implies that the function class
$\Fcal$ is effectively finite: restricted to the $2n$ points appearing
on the right hand side of \eq{slt-eq:symmetrization}, it has \emph{at
  most} $2^{2n}$ elements.  This is because only the values of the
functions \emph{on the sample points and the ghost sample points}
count. The number of effectively different functions can be smaller
than $2^{2n}$, however. For example, it could be the case that $\Fcal$
does not contain a single function which takes value $+1$ on the first
training point. We now want to formalize this.  Let $Z_{n}:=\left(
  (X_1,Y_1),\dots,(X_{n},Y_{n})\right)$ be a given sample of size
$n$.  Denote by
$|\Fcal_{Z_{n}}|$ be the cardinality of $\Fcal$ when restricted to
$\{X_1,\dots,X_{n}\}$, that is, the number of functions from $\Fcal$
that can be distinguished from their values on $\{X_1,\dots,X_{n}\}$.
Let us, moreover, denote the maximum number of functions that can be
distinguished in this way as $\Ncal(\Fcal,n)$, where the maximum runs
over all possible choices of samples, that is

\ba
\Ncal(\Fcal,n) 
= 
\max \{ |\Fcal_{Z_{n}}| \; \big| \; X_1, ..., X_{n} \in \Xcal \}.
\ea

The quantity $\Ncal(\Fcal,n)$ is referred to as the \emph{shattering
coefficient of the function class $\Fcal$ with respect to sample size
$n$}. 
It has a particularly simple interpretation: it is the number of
different outputs $(Y_1,\dots,Y_n)$ that the functions in $\Fcal$ can
achieve on samples of a given size $n$. %
In other words, it measures the \emph{number of ways that the function space can
  separate the patterns into two classes}.  Whenever
$\Ncal(\Fcal,n)=2^{n}$, this means that there exists a sample of size
$n$ on which all possible separations can be achieved by
functions of the class $\Fcal$.  If this is the case, the function space is
said to \emph{shatter} $n$ points. Note that because of the maximum in
the definition of $\Ncal(\Fcal,n)$, shattering means that there
\emph{exists} a sample of $n$ patterns which can be separated in all
possible ways --- it does not mean that this applies to \emph{all}
possible samples of $n$ patterns.\\

The shattering coefficient is a capacity measure of a function class,
that is it measures the ``size'' of a function class in some
particular way. Note that if a function class $\Fcal$ contains very
many functions, then the shattering coefficient tends to be larger
than for a function class which only contains very few
functions. However, the shattering coefficient is more subtle than
simply counting the number of functions in a class. It only counts the
number of functions in relation to the samples we are interested
in. The following section will now finally show how to use the
shattering coefficient to derive a generalization bound for empirical
risk minimization on infinite function classes $\Fcal$. \\

\subsection{Uniform convergence bounds} \label{subsec-uniform-convergence}
Given an arbitrary, possibly infinite function class, we now want to
take a look at the right hand side of 
\eq{slt-eq:symmetrization}.
We now consider a sample of $2n$ points, that is a set $Z_{2n}$, where
we interpret the first $n$ points as the original sample and the
second $n$ points as the ghost sample. The idea is now to replace the
supremum over $\Fcal$ by the supremum over the set $\Fcal_{Z_{2n}}$,
use that the set $\Fcal_{Z_{2n}}$ contains at most $\Ncal(\Fcal,n)
\leq 2^{2n}$ different functions, then apply the union bound on this
finite function set and then the Chernoff bound.  This leads to a
bound like \eq{eq-union-bound}, with $\Ncal(\Fcal,2n)$ playing the role of
$m$. Essentially, those steps can be written down as follows: 

\ba
&\Pr(\sup_{f\in\Fcal} |R(f)-R_\emp(f)|>\epsilon) \\
& \note{due to symmetrization}\\
& 
\leq 2 \Pr(\sup_{f\in\Fcal} |R_\emp(f)-R_\emp'(f)|>\epsilon/2)\\
&\note{only functions in $\Fcal_{Z_{2n}}$ are important}\\
& = 2 \Pr(\sup_{f\in\Fcal_{Z_{2n}}} |R_\emp(f)-R_\emp'(f))|>\epsilon/2)\\
& \note{$\Fcal_{Z_{2n}}$ contains at most $\Ncal(\Fcal, 2n)$
  functions, independently of $Z_{2n}$}\\
&\note{use union bound argument and Chernoff} \\
& \leq 2 \Ncal(\Fcal, 2n) \exp(- n \eps^2/4)
\ea

So all in all we see that 

\banum \label{eq-shattering-bound}
\Pr(\sup_{f\in\Fcal} |R(f)-R_\emp(f)|>\epsilon)
\leq 
2 \Ncal(\Fcal, 2n) \exp(- n \eps^2/4).
\eanum

Now we can use this expression to draw conclusions about consistency
of empirical risk minimization. Namely, ERM is consistent for function
class $\Fcal$ if the right hand side of this expression converges to 0
as $n \to \infty$. Let us look at a few examples. \\

First of all,  consider a case where the shattering coefficient
$\Ncal(\Fcal,2n)$ is  considerably smaller than $2^{2n}$, say 
$\Ncal(\Fcal,2n) \leq (2n)^k$ for some constant $k$ (this means that the
shattering coefficient grows polynomially in
$n$).  Plugging this in the right hand side of
\eqref{eq-shattering-bound}, we get  

\ba
2 \Ncal(\Fcal, 2n) \exp(- n \eps^2/4)
= 
2  \cdot (2n)^k \cdot \exp(- n \eps^2/4)
= 
2 \exp( k \cdot \log(2n) - n \eps^2/4). 
\ea

Here we can see that for all $\eps$, if $n$ is large enough  the whole expression
converges to 0 for $n \to \infty$. From this we can conclude that
whenever the shattering coefficient $\Ncal(\Fcal, 2n)$ only grows
polynomially with $n$, then empirical risk minimization is consistent
with respect to $\Fcal$. \\

On the other hand,
consider the case where we use the function class
$\Fcal_{all}$. It is clear that this class can classify each sample in
every possible way, that is $\Ncal(\Fcal, 2n) = 2^{2n}$ for all values
of $n$. Plugging this in the right hand side of \eqref{eq-shattering-bound}, we get 
\ba
2 \Ncal(\Fcal, 2n) \exp(- n \eps^2/4)
= 
2 \cdot 2^{2n} \exp(- n \eps^2/4)
= 
2 \exp( n ( 2 \log(2) - \eps^2/4 ) ).
\ea

We can immediately see that this expression does not tend to 0 when $n
\to \infty$, that is we cannot conclude consistency for
$\Fcal_{all}$. Note that we cannot directly conclude that ERM using
$\Fcal_{all}$ is inconsistent, either. The reason is that \eqref{eq-shattering-bound}
only gives an upper bound on $\Pr(\sup_{f\in\Fcal}
|R(f)-R_\emp(f)|>\epsilon)$, that is it only provides a sufficient
condition for consistency, not a necessary one. However, with more
effort one can also
prove necessary statements. For example, a necessary and sufficient
condition for consistency of ERM is that 

\ba
\log \Ncal(\Fcal, n)  /n \to 0
\ea

(cf. \citealp{Mendelson03}, for related theorems also see
\citealp{VapChe71,VapChe81}, or Section 12.4. of
\citealp{DevGyoLug96}). 
The proof that this condition is
necessary is more technical, and we omit it. In case of the examples
above, this condition immediately gives the desired results: if
$\Ncal(\Fcal,n)$ is polynomial, then $\log \Ncal(\Fcal,n) /n \to
0$. On the other hand, for $\Fcal_{all}$ 
we always have $\Ncal(\Fcal,n) = 2^{n}$, thus $\log \Ncal(\Fcal,n) /n
= n/n =1$, which does not converge to
0. Thus, ERM using $\Fcal_{all}$ is not consistent. \\

\subsection{Generalization bounds} \label{subsec-generalization-bounds}
It is sometimes useful to rewrite \eq{eq-shattering-bound} ``the other
way round''. That is, instead of fixing $\eps$ and then computing the
probability that the empirical risk deviates from the true risk by
more than $\eps$, we specify the probability with which we want the bound to hold, and then
get a statement which tells us how close we can expect the risk to be
to the empirical risk. This can be achieved by setting the right
hand side of \eq{eq-shattering-bound} equal to some $\delta>0$, and then
solving for $\epsilon$.  As a result, we get the statement 
that with a
probability at least $1-\delta$, any function $f \in \Fcal$ satisfies 
\begin{equation}
  \label{slt-eq:unibound2}
R(f) \le R_\emp(f) + 
\sqrt{\frac{4}{n} \big( \log(2 \Ncal(\Fcal, n)) - \log (\delta)   \big)}.
\end{equation}\index{risk!bound}

In the same way as above, we can use this bound to derive consistency
statements. For example, it is now obvious that empirical risk
minimization is consistent for function class $\Fcal$ if the term
$\sqrt{ \log(2\Ncal(\Fcal,2n)) / n }$ converges to 0 as $n \to
\infty$. Again, this is for example the case if $\Ncal(\Fcal, 2n)$
only grows polynomially with $n$. \\

Note that the bound \eq{slt-eq:unibound2} holds for all functions $f
\in \Fcal$. On the one side, this is a strength of the bound, as it
holds in particular for the function $f_n$ minimizing the empirical
risk, which is what we wanted. Moreover, many learning machines do not
truly minimize the empirical risk, and the bound thus holds for them,
too.  However, one can also interpret it as a weakness, since by
taking into account more information about the function we are
interested in,
one could hope to get more accurate bounds. \\

Let us try to get an intuitive understanding of this bound. It tells
us that if both $R_\emp(f)$ and the square root term are small
simultaneously, then we can guarantee that with high probability, the
risk (i.e., the error on future points that we have not seen yet) will
be small. This sounds like a surprising statement, however, it does
not claim anything impossible. If we use a function class with
relatively small $\Ncal(\Fcal, n)$, i.e., a function class which
cannot ``explain'' many possible functions, and then notice that using
a function of this class we can nevertheless explain data sampled from
the problem at hand, then it is likely that this is not a coincidence,
and we can have captured some essential aspects of the problem.  If,
on the other hand, the problem is too hard to learn from the given
amount of data then we will find that in order to explain the data
(i.e., to achieve a small $R_\emp(f)$), we need a function class which
is so large that it can basically explain anything, and thus the
square root term would be large.  Note, finally, that whether a
problem is hard to learn is entirely determined by whether we can come
up with a suitable function class, and thus by our prior knowledge of
it. Even if the optimal function is subjectively very complex, if our
function class contains that function, and few or no other functions,
we are in an excellent position to learn.

There exists a large number of bounds similar to \eq{eq-shattering-bound}
and its alternative form \eq{slt-eq:unibound2}. Differences occur in
the constants, both in front of the exponential and in its exponent.
The bounds also differ in the exponent of $\epsilon$ (see
\citealp{DevGyoLug96,Vapnik98} and references therein) and in the way
they measure capacity. We will not elaborate on those issues. \\

\subsection{The VC dimension} \label{subsec-vc-dim}
Above we have formulated the generalization bounds in terms of the
shattering coefficient $\Ncal(\Fcal,n)$. The downside is that
they are usually difficult to evaluate. However, there exists a large
number of different capacity concepts, with different advantages and
disadvantages. We now want to introduce the most well known one, the
so-called VC dimension (named after Vapnik and Chervonenkis). Its main purpose is to characterize the
growth behavior of the shattering coefficient using a single number.  \\

We say that a sample $Z_n$ of size $n$ is {\em shattered by function class }
$\Fcal$ if the function class can realize any labeling on the given
sample, that is $|\Fcal_{Z_n}| = 2^n$. The {\em VC dimension of
  $\Fcal$}, denoted by $\VC(\Fcal)$, is now defined as the largest
number $n$ such that there exists a sample of size $n$ which is
shattered by $\Fcal$. Formally, 
\ba
\VC(\Fcal)
 = \max \{n \in \Nat \condon | \Fcal_{Z_n}| = 2^n \text{ for some } 
Z_n \}.
\ea
If the maximum does not exist, the VC dimension is defined to be
infinity. For many
examples of function classes and their VC dimensions see for example
\citet{KerVaz94} or \citet{AntBig92}. \\

A beautiful combinatorial result proved simultaneously by several
people (\citealp{Sauer72}, \citealp{Shelah72}, \citealp{VapChe71}) characterizes the growth behavior of the
shattering coefficient and relates it to the VC dimension: 

\begin{lemma}[Vapnik, Chervonenkis, Sauer, Shelah]
Let $\Fcal$ be a function class with finite VC dimension $d$. Then 
\ba
\Ncal(\Fcal,n) \leq \sum_{i=0}^d \binom{n}{i}
\ea
for all $n \in \Nat$. In particular, for all $n \geq d$ we have 
\ba
\Ncal(\Fcal,n) \leq \left(\frac{en}{d}\right)^d.
\ea
\end{lemma}

The importance of this statement lies in the last fact. If $n \geq d$,
the shattering coefficient behaves like a polynomial function of the
sample size $n$. This is a very remarkable result: once we know the
VC-dimension of a function class $\Fcal$ is finite, we already know
that the shattering coefficients grow polynomially with $n$. By the
results of the last section this implies the consistency of ERM. Note
that we also have a statement in the other direction. If the
VC-dimension is infinite, this means that for each $n$ there exists
some sample which  can be shattered by $\Fcal$, that is
$\Ncal(\Fcal,n) = 2^n$. For this case we have already seen above that
ERM is not consistent. Together, we achieve the following important
characterization: 

\begin{theorem}
Empirical risk minimization is consistent with respect to $\Fcal$ if
and only if 
$\VC(\Fcal)$ is finite. 
\end{theorem}

One important property to note both about the shattering coefficient
and the VC dimensions is that they do not depend on the underlying
distribution $P$, they only depend on the function class $\Fcal$. One
the one hand, this is an advantage, as all generalization bounds
derived from those concepts apply to all possible probability
distributions. On the other hand, one can also consider this as
disadvantage, as the capacity concepts do not take into account
particular properties of the distribution at hand. In this sense,
those capacity concepts often lead to rather loose bounds. \\

\subsection{Rademacher complexity} \label{subsec-rademaacher}

A different concept to measure the capacity of a function space is
the Rademacher complexity. Compared to the shattering coefficient and
the VC dimension, it does depend on the underlying probability
distribution, and usually leads to much sharper bounds than both of
them. The Rademacher complexity is defined as follows. Let $\sigma_1,
\sigma_2, ...$ independent random variables which attain the two values
$+1$ and $-1$ with probability $0.5$ each (such random variables are
sometimes called Rademacher variables, therefore the name ``Rademacher
complexity''). For example, they could be
the results of repeatedly tossing a fair coin. We formally define the
Rademacher complexity $\Rcal(\Fcal)$ of a function space $\Fcal$ as

\banum \label{eq-rademacher}
\Rcal(\Fcal) := E \sup_{f \in \Fcal} \frac{1}{n} \sum_{i=1}^n \sigma_i
f(X_i).
\eanum

This expression looks complicated, but it has a nice
interpretation. For the moment, consider the values $\sigma_i$ as
fixed, and interpret label $\sigma_i$ as a label of the point
$X_i$. As both $\sigma_i$ and $f(X_i)$ only take the values $+1$ or
$-1$, the product $\sigma_i f(X_i)$ takes the value $+1$ if $\sigma_i = f(X_i)$, and 
$-1$ if $\sigma_i \neq f(X_i)$. As a consequence, the sum on the right
hand side of Equation \eqref{eq-rademacher} will be large if the
labels $f(X_i)$ coincide with the
labels $\sigma_i$ on many data points. This means that the function
$f$ ``fits well'' to the labels $\sigma_i$:  if the
labels $\sigma_i$ were the correct labels, $f$ would have a small
training error $\Remp$. Now taking into account the supremum, we not
only look at one function $f$, but at all functions $f \in \Fcal$. We
can see that $ \sup_{f \in \Fcal} \sum_{i=1}^n \sigma_i f(X_i)$ is
large if there exists a function in $\Fcal$ which fits well to the
given sequence $\sigma_i$ of labels. Finally, recall that the labels
$\sigma_i$ are supposed to be random variables. We can consider them
as  random labels on the data
points $X_i$. As we take the expectation over both the data points and
the random labels, the overall Rademacher complexity is high if the
function space $\Fcal$ is able to ``fit well'' to random labels. This
intuition makes sense: a function space has to be pretty large to be
able to fit to all kinds of random labels on all kinds of data
sets. In this sense, the Rademacher complexity measures how
``complex'' the function space is: the higher $\Rcal(\Fcal)$, the
larger the complexity of $\Fcal$. \\

From a mathematical point of view, the Rademacher complexity is
convenient to work with. One can prove generalization bounds of the
following form: with probability at least $1-\delta$, 

\ba
R(f) \leq \Remp(f) + 2 \Rcal(\Fcal) + \sqrt{\frac{\log (1/\delta)}{2n}}
\ea

Rademacher complexities have some advantages over the classical
capacity concepts such as the VC dimension. Most notably, the bounds
obtained by Rademacher complexities tend to be much sharper than the
ones obtained by the classical tools.  The proof techniques are
different from the ones explained above, but we do not want to go into
details here. For literature on Rademacher complexity bounds, see for
example \citet{Mendelson03}, \citet{BouBouLug03} or
\citet{BouBouLug05}
and references therein.  \\

\subsection{Large margin bounds} \label{subsec-large-margin}

\begin{figure}[t]
  \centerline{\includegraphics[width=0.4\textwidth]{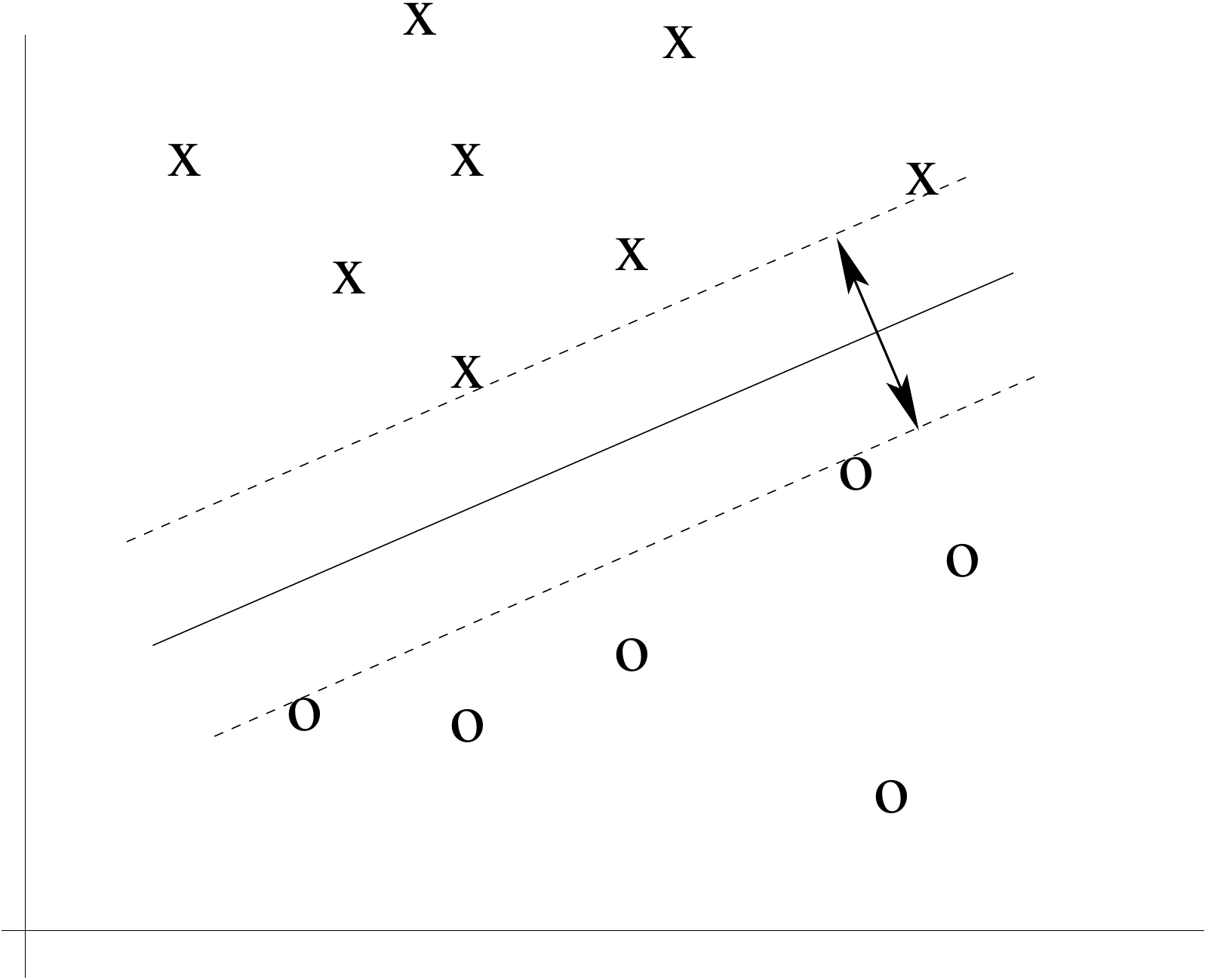}}
\caption{\em Margin of a linear classifier: the crosses depict training
  points with label +1, the circles training points with label -1. The
  straight line is the linear classifier $f_n$, and the dashed line
  shows the margin. The width $\rho$ of the margin is depicted by the arrow.
\label{fig-margin}
}
\end{figure}

Finally, we would like to introduce another type of capacity measure
of function classes which is more specialized than the general
combinatorial quantities introduced above. Consider the special case
where the data space consists of points in the two-dimensional space
$\R^2$, and where we want to separate classes by a straight line. Given a
set of training points and a classifier $f_n$ which can perfectly
separate them, we define the {\em
  margin of the classifier $f_n$} as the smallest distance of any
training point to the separating line $f_n$
(cf. Figure~\ref{fig-margin}). Similarly, a margin can be defined for
general linear classifiers in the space $\R^d$ of arbitrary dimension $d$. 
 It can be proved that the VC
dimension of a class $\Fcal_{\rho}$ of linear functions with all have margin at least $\rho$ can
essentially be
bounded by the ratio of the radius $R$ of the smallest sphere enclosing
the data points with the margin $\rho$, that is 
\ba
\VC(\Fcal_{\rho}) \leq \min\left\{ d, \frac{4 R^2}{\rho^2} \right\} +1.
\ea
(cf. Theorem 5.1. in \citealp{Vapnik95}). 
That is, the larger the margin $\rho$ of the functions in the class
$\Fcal_{\rho}$, the smaller is its VC dimension. Thus, one can use the
margin of classifiers as a capacity concept. One of
the most well-known classifiers, the support vector machine (SVM)
builds on this result. See \citet{SchSmo02} for a comprehensive
treatment. An example of a generalization bound involving the large
margin is as follows (for a more precise statement see for example
Theorem 7.3. in \citealp{SchSmo02}):

\begin{theorem}[Large margin bound] \label{th-large-margin}
Assume the data space lies inside a ball of radius $R$ in
$\R^d$. Consider the set $\Fcal_{\rho}$ of linear classifiers with margin at
least $\rho$. Assume we are given $n$ training examples. By $\nu(f)$
denote the fraction of training examples with margin
smaller than $\rho$ or which are on the wrongly classified by 
some classifier $f \in \Fcal_{\rho}$. Then, with probability at least $1 - \delta$, the true 
error of any $f \in \Fcal_{\rho}$ can be bounded by 

\ba
R(f) \leq \nu(f) + \sqrt{\frac{c}{n} \left( \frac{R^2}{\rho^2} \log(n)^2
    + \log(1/\delta)\right)}
\ea

where $c$ is some universal constant.

\end{theorem}
\subsection{Other generalization bounds and capacity concepts}

Above we have introduced a few capacity concepts for function classes
such as the shattering coefficient, the VC dimension, or the
Rademacher complexity. In the literature, there exist many more
capacity concepts, and introducing all of them will be beyond the
scope of this overview. However, we would like to point out the
general form which most generalization bounds take. Usually, those
bounds are composed of three different terms and have a form like:
with probability at least $1 - \delta$, 

\ba
R(f) \leq \Remp(f) + capacity(\Fcal) + confidence(\delta). 
\ea

That is, one can bound the true risk of a classifier by its empirical
risk, some capacity term which in the simplest case only depends on
the underlying function class, and a confidence term which depends on
the probability with which the bound should hold. Note that by nature,
all bounds of this form a worst case bounds: as the bound holds for
all functions in the class $\Fcal$, the behavior of the bound is
governed by the ``worst'' or ``most badly behaved'' function in the
class. This point is often used to criticize this approach to
statistical learning theory, as natural classifiers do not tend to
pick the worst function in a class.

\section{Incorporating knowledge into the bounds } \label{sec-prior-knowledge}

In the previous section we presented the standard approach to derive
risk bounds using uniform laws of large numbers.  The bounds are agnostic
in the sense that they do not make any prior assumptions on the
underlying distribution $P$. \\

In many cases, however, it is desirable to be able to take into
account some prior knowledge we might have about our problem. There
are several good reasons for doing so. First, the bounds we considered
above are worst case bounds over all possible distributions. That is,
their behavior might be governed by completely unnatural distributions
which would never occur in practice (such as distributions on Cantor
sets, for example). In this sense, the bounds are overly
pessimistic. As we often believe that ``real'' distributions have
certain regularity aspects, we might improve the results obtained
above by making use of this regularity. A more fundamental reason for
incorporating prior knowledge is the no free lunch theorem, which will
be discussed in Section \ref{sec-nfl}. In a nutshell, the no free
lunch theorem states that learning is impossible unless we make
assumptions on the underlying distribution. Thus it seems a good idea
to work out ways to state prior assumptions on distributions, and to
build them into our learning framework. In this section we want to
point out a few ways of
doing so. \\

\subsection{How to encode prior knowledge in the classical approach}

Let us step back and try to figure out where prior knowledge enters a
learning problem: \\

\begin{enumerate}
\item { Via the formalization of the data space $\Xcal$, for example
  its topology. For example, we might construct a distance function or
  a similarity function which tells us how ``similar'' different input
  values in $\Xcal$ are. A learning algorithm then tries to assign
  similar output labels to similar input labels. The topology of
  $\Xcal$ is one of the most important places to encode prior
  knowledge, not only for theory but in particular  for practical
  applications. } 

\item { Via the elements of $\Fcal$. Here we encode our assumptions on
  how a useful classifier might look. While in theory this is an
  important way to encode prior assumptions, it does not play such a
  big role in practice. The reason is that it is not so easy to
  formulate any prior assumptions in terms of the function space. In
  applications, people thus often use one out of a handful of standard
  function spaces, and do not consider this as the main place to
  encode prior assumptions. An exception are Bayesian approaches, see
  below. }

\item { Via the loss function $\ell$. This function encodes what the
  ``goal'' of learning should be. While the loss functions we
  mentioned above are pretty straightforward, there are many ways to
  come up with more elaborate loss functions. For example, we can
  weight errors on individual data points more heavily than  those on
  other data points. Or we can weight different kinds of errors
  differently. For example, in many problems the ``false positives''
  and ``false negatives'' have different costs associated to them. In
  the example of spam filtering, the cost of accidentally labeling a spam
  email as ``not spam'' is not so high (the user can simply delete
  this mail). However, the cost of labeling a ``non-spam'' email as
  ``spam'' can be very high (the mail might have contained important
  information for the user, but got deleted without his knowledge). 
}

\item {Via assumptions on the underlying probability distributions. So
    far, the classical approach presented above was agnostic in the
    sense that it did not make any assumptions on the underlying
    probability distribution. That is, no matter what probability
    distribution $P$ generated our data, the generalization bounds
    from above apply. Due to this property they are often called
    ``worst case bounds'', as they even hold for the worst
    and most unlikely distribution one can imagine. \\

    This worst case approach has often been criticized as being overly
    pessimistic. The accusation is that the behavior of the results is
    governed by artificial, pathological probability distributions
    which would never occur in practice, and thus the results do not
    have much meaning for the ``real world''. To circumvent this, one
    could make assumptions on the underlying probability
    distributions. For example, one could assume that the distribution
    is ``nice'' and ``smooth'', that the labels are not overly noisy,
    and so on.  Such assumptions can lead to a dramatic improvement of
    the learning guarantees, as is discussed in Section~\ref{subsec-rates} in
    connection with the fast learning rates.  On the other hand, making
    assumptions on the distributions contradicts the agnostic approach
    towards learning we wanted to make in the beginning. It is often
    perceived problematic to make such assumptions, because after all
    one does not know whether those assumptions hold for ``real
    world'' data, and guarantees proved under those assumptions could be
    misleading in cases where the
    assumptions are violated. 
}

\end{enumerate}

In some sense, all those quantities mentioned above do enter the classical
generalization bounds: The loss function is directly evaluated in the
empirical risk of the bound. The topology of $\Xcal$ and the choice of
$\Fcal$ are both taken into account when computing the capacity
measure of $\Fcal$ (for example the margin of functions).  \\

However, the capacity measures are often perceived as a rather
cumbersome and crude way to incorporate such assumptions, and the
question is whether there are any other ways to deal with
knowledge. In this section we now want to focus on a few advanced
techniques, all of which try to improve the framework above from
different points of view. The focus of the attention is on how to
incorporate more ``knowledge'' in the bounds than just counting
functions in $\Fcal$: either knowledge we have a priori, or knowledge
we can obtain a posteriori.

\subsection{PAC Bayesian bounds} \label{subsec-pac-bayesian}

The classical SLT approach has two features which are important to
point out. The capacity term in the generalization bound always
involves some quantity which measures the ``size'' of the function
space $\Fcal$. However, this quantity usually does not directly depend
on the complexity of the individual functions in $\Fcal$, it rather
counts how many functions there are in
$\Fcal$. Moreover, the bounds do not contain any quantity which
measures the complexity of an individual function $f$ itself. In this
sense, all functions in the function space $\Fcal$ are treated the
same: as long as they have the same training error, their bound on the
generalization error is identical. No function is singled out in any
special way. \\

This can be seen as an advantage or a disadvantage. If we believe that
all functions in $\Fcal$ are similarly well suited to fit a certain
problem, then it would not be helpful to introduce any ``ordering''
between them.  However, often this is not the case. We already have some
``prior knowledge'' which we accumulated in the past. This knowledge
might tell us that some functions $f \in \Fcal$ are much more likely
to be a good classifier than others. The Bayesian approach is one way
to try to incorporate such prior knowledge into statistical
inference. The general idea is to introduce some prior distribution
$\pi$ on the function space $\Fcal$. This prior distribution expresses
our belief about how likely a certain function is to be a good
classifier. The larger the
value $\pi(f)$ is, the more confident we are that $f$ might be a good
function. The important point is that this prior will be chosen
before we get access to the data. It should be selected only based on
background information or prior experience. \\

It turns out that this approach can be effectively combined with the
classic SLT framework. For example, one can prove that for a finite function
space $\Fcal$, 
with probability at least $1 -\delta$, 
\ba
R(f) \leq \remp(f) + \sqrt{\frac{\log(1/ \pi(f)) + \log(1/\delta)}{2n}}
\ea
where $\pi(f)$ denotes the value of the prior on $f$.  This is the
simplest PAC-Bayesian bound. The name comes from the fact that it
combines the classical bounds (sometimes called PAC bounds where
``PAC'' stands for ``probably approximately correct'') and the
Bayesian framework. Note that the
right hand side does not involve a capacity term for $\Fcal$, but
instead ``punishes'' individual functions $f$ according to their prior
likelihood $\pi(f)$. The more unlikely we believed $f$ to be, the
smaller $\pi(f)$, and the larger the bound. This mechanism shows that
among two functions with the same empirical risk on the training data,
one prefers the one with the higher prior value $\pi(f)$. 
For background reading on PAC-Bayesian bounds, see for example Section
6 of \citet{BouBouLug05} and references therein. 

\subsection{Luckiness approach}

\begin{figure}[t]
  \begin{center}
    \includegraphics[width=0.6\textwidth]{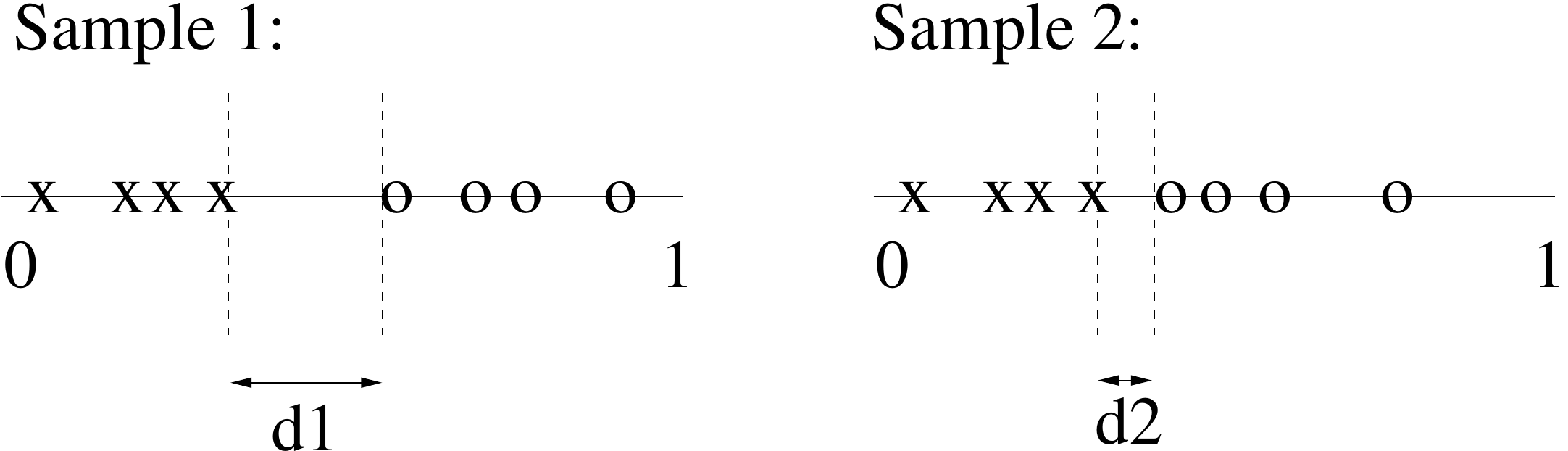}
  \end{center}
\caption{\em Luckiness approach: }
\label{fig-luckiness}
\end{figure}

There is one  important aspect of the bounds obtained in the
classical approach (and in the PAC-Bayesian approach as well): all
quantities in the capacity term of the bound have to be evaluated {\em
  before} getting access to the data. The mathematical reason for this
is that capacity measures are not treated as random quantities, they
are considered to be deterministic functions of a fixed function
class. For example, the large margin bound in Section~\ref{subsec-large-margin} has to be
parsed as follows: ``if we knew that the linear classifier we are going
to construct has margin $\rho$, then the generalization bound of
Theorem~\ref{th-large-margin} would hold''. From this statement, it becomes apparent
that there is a problem: in practice, we do not know in advance
whether a given classifier will have a certain margin, this will
strongly depend on the data at hand.  At first glance, this
problem sounds superficial, and many people in the machine learning
community are not even aware of it. It is closely related to the
question what prior assumptions we want to encode in the learning
problems before seeing actual data, and which properties of a problem
should only be evaluated a posteriori,
that is after seeing the data. While the PAC-Bayesian framework deals
with a priori knowledge, we still need a framework to deal with ``a
posteriori'' knowledge. \\

This problem is tackled in the ``luckiness framework'', first introduced in \citet{ShaBarWilAnt98}
and then considerably extended in \citet{HerWil02}. On a high
level, the idea is as follows. Instead of proving bounds where the
capacity term just depends on the underlying function class, one would
like to allow this capacity to depend on the actual data at hand. Some
samples are considered to be very ``lucky'', while others are
``unlucky''. The rationale is that for some samples it is easy to
decide on a good function from the given function class, while the
same is difficult for some other samples. As a result, one can
prove generalization bounds which do depend on the underlying sample
in terms of their ``luckiness value''. \\

We would like to illustrate the
basic idea with a small toy example. Suppose our data space $\Xcal$ is
just the interval $[0,1]$, and define the functions $f_a$ as 
\ba
f_a(x) = \begin{cases} 
-1 & \text{ if } x \leq a \\ 
+1 & \text{ if } x  > a 
\end{cases}
\ea
As function space we use the space $\Fcal := \{ f_a \condon a \in
[0,1]\}$. Moreover, assume that the true underlying distribution can
be classified by a function in $\Fcal$ with 0 error, that is we do not
have any label noise. Now consider two particular samples, see Figure
\ref{fig-luckiness} for illustration. In the first sample, the two
data points which are closest to the decision boundary have distance
$d_1$, which is fairly large compared to the corresponding distance
$d_2$ in the second sample. It is clear that the true decision
boundary has to lie somewhere in the interval $d_1$ ($d_2$,
respectively). The uncertainty about the position of the decision
boundary is directly related to the lengths $d_1$ and $d_2$,
respectively: we can be confident that the error on Sample 2 will
be much smaller than the error on Sample 1. In this sense, Sample 2 is
more ``lucky'' than Sample 1. Now consider two classifiers $f_1$ and
$f_2$ constructed using Sample 1 and Sample 2, and which have 0
training error. Classic generalization
bounds such as the ones in Section~\ref{sec-capacity} give exactly the same risk value to both
classifiers: they have the same training error and use the same
function space $\Fcal$. 
This is where the luckiness approach comes in. It assigns a
``luckiness value'' to all samples, and then derives a generalization
bound which not only depends on the training error and the capacity of
the underlying function class, but also on the luckiness value of the
sample at hand. The technical details are fairly complicated, and we
refrain from introducing them here. Let us just remark that in the
example above, the luckiness approach would attribute a smaller risk
to Sample 2 than to Sample 1. For more material on how this works
exactly see \citet{ShaBarWilAnt98} and \citet{HerWil02}. Also note
that similar approaches have been developed in classical statistics
under the name ``conditional confidence sets'', see \citet{Kiefer77}.

\section{The approximation error and Bayes consistency} \label{sec-approx-error}

In the previous sections we have investigated the standard approach
to  bound the estimation error of a classifier. This is enough
to achieve consistency with respect to a given function class $\Fcal$. In this section, we now want to look
at the missing piece towards Bayes-consistency: the approximation
error.\\

Recall that the estimation error was defined as $ R(f_n) -
R(f_{\Fcal})$ and the approximation error as $ R(f_{\Fcal}) -
R(f_{Bayes})$, cf. Section~\ref{subsec-bias-variance}. In order to
achieve Bayes-consistency, both terms have to vanish when $n
\to \infty$.  We have seen above that for the estimation error to
converge to 0 we have to make sure that the function space $\Fcal$ has
a reasonably small capacity. But this now poses a problem for the
approximation error: if the function space $\Fcal$ has a small
capacity, this means in particular that the space $\Fcal$ is
considerably smaller than the space $\Fcal_{all}$ of all
functions. Thus, if we fix the function class $\Fcal$ and the Bayes
classifier $f_{Bayes}$ is not contained
in $\Fcal$, then the approximation error might not be 0. \\

There are only two ways to solve this problem. The first one is to
make assumptions on the functional form of the Bayes classifier. If $f_{Bayes} \in \Fcal$ for some known function space $\Fcal$ with
small capacity, we know that the approximation error is 0. In this
case, Bayes-consistency reduces to consistency with respect to $\Fcal$, which can be
achieved by the methods discussed in the last section. However, if we
do not want to make assumptions on the Bayes classifier, then we have
to choose a different construction. \\

\subsection{Working with nested function spaces}

In this construction, we will not only consider one function class
$\Fcal$, but a sequence $\Fcal_1, \Fcal_2, ... $ of function
spaces. When constructing a classifier on $n$ data points, we will do
this based on function space $\Fcal_n$. The trick is now that the
space $\Fcal_n$ should become more complex the larger the sample size
$n$ is. The standard construction is to choose the spaces $\Fcal_n$
such that they form an increasing sequence of nested function spaces,
that is $\Fcal_1 \subset \Fcal_2 \subset \Fcal_3 \subset ... $. The
intuition is that we start with a simple function space and then
slowly add more and more complex functions to the space. If we are now
given a sample of $n$ data points, we are going to pick our classifier
from the space $\Fcal_n$. If we want this classifier to be
Bayes-consistent, there are two things we need to ensure:

\begin{enumerate}
\item The estimation error has to converge to 0 as $n \to \infty$. To
  this end, note that for each fixed $n$ we can bound the estimation
  error by one of the methods of Section~\ref{sec-capacity}. This
  bound is decreasing as the sample size increases, but it is
  increasing as the complexity term increases. We now have to make
  sure that the overall estimation error is still decreasing, that is
  the complexity term must not dominate the sample size term. To
  ensure this, we have to make sure that the complexity of $\Fcal_n$
  does not grow too fast as the sample size increases.

\item The approximation error has to converge to 0 as $n \to
  \infty$. To this end, we need to ensure that eventually for some
  large $n$, each
  function of $\Fcal_{all}$ is either contained in $\Fcal_n$, or that
  it can be approximated by a function from $\Fcal_n$. We are 
  going to discuss below how this can be achieved. 

\end{enumerate}

An example how those two points can be stated in a formal way is the
following theorem, adapted from Theorem 18.1 of \citet{DevGyoLug96}:

\begin{theorem}
Let $\Fcal_1, \Fcal_2, ...$ be a sequence of function spaces, and
consider the classifiers 
\ba
f_n = \argmin_{f \in \Fcal_n} \remp(f).
\ea
Assume that for any distribution $P$ the following two conditions are
satisfied: 
\begin{enumerate}
\item The VC-dimensions of the spaces $\Fcal_n$ satisfy 
$ VC(\Fcal_n) \cdot \log n / n \to 0$ as $n \to \infty$. 
\item $R(f_{\Fcal_n}) \to R(f_{Bayes})$ as $n \to \infty$.\\
\end{enumerate}
Then, the sequence of classifiers $f_n$ is Bayes-consistent. 
\end{theorem}

Let us try to understand this theorem. We are given a sequence of
increasing function spaces $\Fcal_n$. For each sample of size $n$,
we pick the function in $\Fcal_n$ which has the lowest empirical
risk. This is our classifier $f_n$. If we want this to be consistent,
two conditions have to be satisfied. The first condition
says that the complexity of the function classes, as measured by the
VC dimension, has to grow slowly. For example, if we choose the
function spaces $\Fcal_n$ such that $VC(\Fcal_N) \approx n^{\alpha}$
for some $\alpha \in ]0,1[$, then
the first condition is satisfied because 
\ba
 VC(\Fcal_n) \cdot (\log n) / n
\;\approx\; 
n^\alpha (\log n) /n  \;= \;(\log n) / n^{1 - \alpha} \;\to\; 0.
\ea
However, if we choose $VC(\Fcal_N)
\approx n$ (that is, $\alpha=1$ in the above calculation), then this is no longer the case: 
\ba
 VC(\Fcal_n) \cdot (\log n) / n
\approx 
\log n \to \infty.
\ea
The second condition of the theorem simply states that the
approximation error has to converge to 0, but the theorem does not
give any insight how to achieve this. But as we discussed above it is
clear that the latter can only be achieved for an increasing sequence
of function classes.

\subsection{Regularization}

An implicit way of working with nested function spaces is the
principle of regularization. Instead of minimizing the empirical risk
$\remp(f)$ and then expressing the generalization ability of the
resulting classifier $f_n$ using some capacity measure of the
underlying function class $\Fcal$, one
can pursue a more direct approach: one directly minimizes a the
so-called regularized risk 

\ba
\rreg(f) = \remp(f) + \lambda \Omega(f). 
\ea

Here, $\Omega(f)$ is the so-called regularizer. This regularizer is
supposed to punish overly complex functions. For example, one often
chooses a regularizer which punishes functions with large
fluctuations, that is one chooses $\Omega(f)$ such that it is small
for functions which vary slowly, and large for functions which
fluctuate a lot. Or, as an other example, for linear classifiers one can choose $\Omega(f)$ as
the inverse of the margin of a function (recall the definition of a
margin in Section \ref{subsec-large-margin}). \\

The $\lambda$ in the definition of the regularized
risk is a trade-off constant. It ``negotiates''  between the
importance of $\remp(f)$ and of $\Omega(f)$. If $\lambda$ is very
large, we take the punishment induced by $\Omega(f)$ very seriously,
and might prefer functions with small $\Omega(f)$ even if they have a
high empirical risk. On the other hand, if $\lambda$ is small, the
influence of the punishment decreases, and we merely choose functions
based on their empirical risks. \\

The principle of regularization consists in choosing the classifier
$f_n$ that minimizes the regularized risk $\rreg$. Many of the
widely-used classifiers can be cast into the framework of
regularization, for example the support vector machine (see
\citealp{SchSmo02}, for details). \\

To prove Bayes-consistency of regularized classifiers one essentially
proceeds as outlined in the subsection above: for some slowly increasing
sequence $\omega_1, \omega_2, ...$ we consider nested
function spaces $\Fcal_{\omega_1}$, $\Fcal_{\omega_2}$, ... , where each
$\Fcal_{\omega_i}$ contains all functions $f$ with $\Omega(f) \leq
\omega_i$. Eventually, if $i$ is very large, the space
$\Fcal_{\omega_i}$ will approximate the space $\Fcal_{all}$ of all
functions. For consistency, one  has to take the constant
$\lambda$ to 0 as $n \to \infty$. This ensures that eventually, for
large $n$ we indeed are allowed to pick functions from a space close
to $\Fcal_{all}$. On the other hand, the constant $\lambda$ must not
converge to 0 too fast, otherwise we will already start overfitting
for small values of $n$ (as with a small constant $\lambda$, one
essentially ignores the regularizer and consequently performs
something close to ERM over a very large
set of functions). A paper which carefully goes through all those steps for the
example of the support vector machine is \citet{Steinwart05}. \\

Note that there is one important conceptual difference between empirical risk
minimization and regularization. In regularization, we have a function
$\Omega$ which measures the ``complexity'' of an individual function
$f$. In ERM, on the other side, we never look at complexities of
individual functions, but only at the complexity of a function
class. The latter, however, is more a measure of capacity, that is a
measure of the ``number of functions'' in $\Fcal$, and only indirectly
a measure of how complex the individual functions in the class
are. From an intuitive point of view, the first approach is often
easier to grasp, as the complexity of an individual function is a more
intuitive concept than the capacity of a function class.

\subsection{Achieving zero approximation error}

The theorem above shows the general principle how we can achieve
Bayes-consistency. However, the theorem simply postulated as its
second condition that the approximation error should converge to
0. How can this be achieved in practice?
It turns out that there are many situations where this is not so
difficult. Essentially we have to make sure that each function of
$\Fcal_{all}$ is either contained in $\Fcal_n$ for some large $n$, or that it can be
approximated arbitrarily well by a function from $\Fcal_n$. The area
in mathematics dealing with such kind of problems is called
approximation theory, but for learning theory purposes simple
approximation results are often enough (for more sophisticated ones see
for example \citet{CuckerZhou07}). The only technical problem we have
to solve is that we need a statement of the following form: if two
functions are ``close'' to each other, then their corresponding risk
values are also ``close''. \\

Such statements are often quite easy to obtain.  For example, it is
straightforward to see that if $f$ is a binary classification
function (i.e., with $f(x) \in \{\pm 1\}$) and $g$ any arbitrary
(measurable) function, and the $L_1$-distance between $f$ and $g$ is less
than $\delta$, then so is their difference in
0-1-risk, i.e., $P( f(x) \neq \sgn(g(x)) )
< \delta$. This means that in order to prove that the
approximation error of a function space $\Fcal$ is smaller than
$\delta$, we just have to know that every function in $\Fcal_{all}$
can be approximated up to $\delta$ in the $L_1$-norm by functions from
$\Fcal$. 
Results of this kind are abundant in the mathematics
literature. For example, if $\Xcal$ is a bounded subset of the real
numbers, it is well known that one can approximate any measurable
function on this set arbitrarily well by a polynomial. Hence, we could
choose the spaces $\Fcal_n$ as spaces of polynomials with degree at
most $d_n$ where $d_n$ slowly grows with $n$.  This is enough to
guarantee convergence of the approximation error. \\

\subsection{Rates of convergence} \label{subsec-rates}

We would like to point out one more fundamental difference between
estimation and approximation error: the uniform rates of
convergence. In a nutshell, a ``rate of convergence'' gives
information about how ``fast'' a quantity converges. In our case, the
rate says something about how large $n$ has to be in order to ensure
that an error is smaller than a certain quantity. Such a statement
could be: in order to guarantee (with high probability) that the
estimation error is smaller than 0.1 we need at least $n=1000$ sample
points. Usually it is the case that rates of convergence depend on
many parameters and quantities of the underlying problem. In the
learning setting, a particularly important question is whether the
rate of convergence also depends on the underlying distribution. And
this is where the difference between estimation error and
approximation error happens. For the estimation error, there exist
rates of convergence which hold independently of the underlying
distribution $P$. This is important, as it tells us that we can
give convergence guarantees even though we do not know the underlying
distribution. %
For the approximation error, however, it is not possible to give rates
of convergence which hold for all probability distributions $P$. This
means that a statement like ``with the nested function classes
$(\Fcal_n)_n$ we need at least $n=1000$ to achieve approximation error
smaller than 0.01'' could only be made if we knew the underlying
probability distribution. One can prove that unless one makes some
further assumptions on the true distribution, for any fixed sequence
$(\Fcal_n)_n$ of nested function spaces the rate of convergence of the
approximation error can be arbitrarily slow. Even though we might know
that it eventually converges to 0 and we obtain consistency, there is
no way we could specify what ``eventually''
really means.\\

It is important to note that statements about rates of convergence
strongly depend on the underlying assumptions. For example, above we
have already pointed out that even under the assumption of independent
sampling, no uniform rate of convergence for the approximation
error exists.  A similar situation occurs for the estimation error if
we weaken the sampling assumptions. If we no longer require that
samples $X_i$ have to be independent from each other, then the
situation changes fundamentally. While for ``nearly independent
sampling'' (such as stationary $\alpha$-mixing processes) it might
still be possible to recover results similar to the ones presented
above, as soon as we leave this regime is can become impossible to
achieve such results.  (For the specialists: even in the case
of stationary ergodicity, we can no longer achieve universal consistency, see \citealp{Nobel99}. ) 
For more discussion see also \citet{SteHusSco06} and references therein. \\

On the other hand, if we strengthen our sampling assumptions, we can
even improve the rates of convergence. If we assume that the data
points are sampled independently, and if we make a few assumptions on
the underlying distribution (in particular on the label noise), then
the uniform rates of convergence of the estimation error can be
improved dramatically. There is a whole branch of learning theory
which deals with this phenomenon, usually called ``fast rates''. For
an overview see Section 5.2. of
\citet{BouBouLug05}. \\

Finally, note that both consistency and rates of convergence deal with
the behavior of an algorithm as the sample size tends to
infinity. Intuitively, consistency is a ``worst case statement'': it
says that ultimately, an algorithm will give a correct solution and
does not make systematic errors. Rates
of convergence, on the other hand, make statements about how ``well
behaved'' an algorithm can be. Depending on prior assumptions, one can
compare different algorithms in terms of their rates of
convergence. This can lead to insights into which learning algorithms might
be better suited in which situations, and it might help us to choose a
particular algorithm for a certain application on which we
have prior knowledge.

\section{No free lunch theorem} \label{sec-nfl}

\begin{figure}[t]
  \begin{center}
    \includegraphics[width=0.6\textwidth]{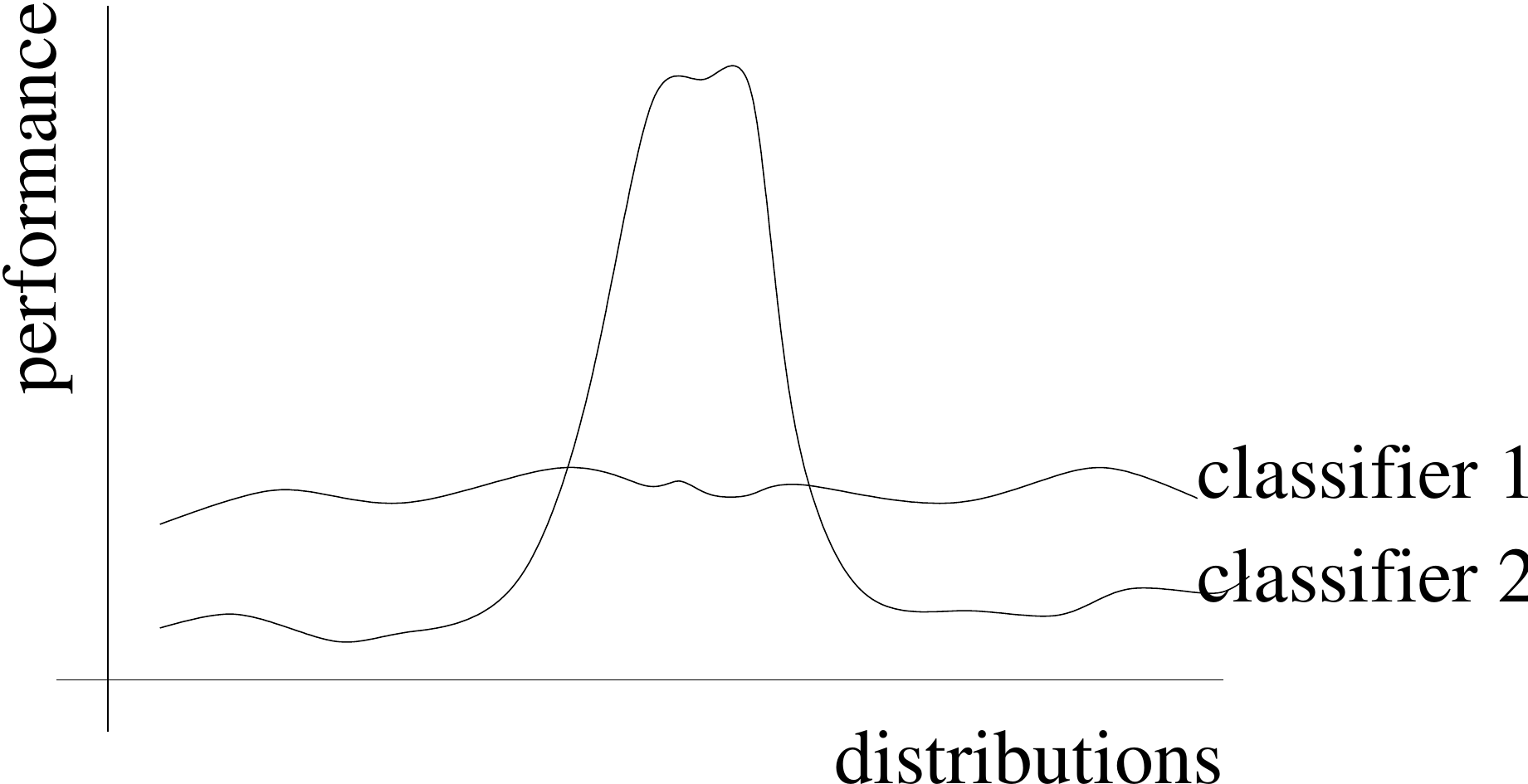}
  \end{center}
  \caption{\em No free lunch theorem: Classifier 1 depicts a general
    purpose classifier. It performs moderately on all kinds of
    distributions. Classifier 2 depicts a more specialized classifier
    which has been tailored towards particular distributions. It
    behaves very good on those distributions, but worse on other
    distributions. According to the no free lunch theorem, the average
    performance of all classifiers over all distributions, that is the
    area under the curves, is identical for all classifiers. }
\label{fig-nfl}
\end{figure}

Up to now, we have reported many positive results in statistical
learning theory. We have formalized the learning problem, defined
the goal of learning (to minimize the risk), specified which properties a
classifier should have (consistency), and devised a framework to
investigate these properties in a fundamental way. Moreover, we have
seen that there exist different ways to achieve consistent classifiers
($\k$-nearest neighbor, empirical risk minimization), and of course
there exist many more consistent ways to achieve consistency. Now it is a natural
question to ask which of the consistent classifiers is ``the best''
classifier.  \\

Let us first try to rephrase our question in a more formal way. Under
the assumption of independent sampling from some underlying, but unknown
probability distribution, is there a classifier which ``on average
over all probability distributions'' achieves better results than any
other classifier? Can we compare classifiers pairwise, that is compare
whether a classifier A is better than classifier B, on average over
all distributions? The reasons why we consider statements ``on average
over all distributions'' lies in the fact that we do not want to make
any assumption on the underlying distribution. Thus it seems natural
to study the behavior of classifiers on any possible distribution. 
Regrettable, those questions have  negative answers
which are usually stated in the form of the so-called ``no free lunch
theorem''. A general proof of this theorem appeared in
\citet{WolMac97} and \citet{Wolpert01}.  A simpler but more accessible
version for finite data spaces has been published  by \citet{HoPep02}.
For versions with focus on convergence rates, see Section 7 of \citet{DevGyoLug96}.\\

For the ease of understanding, consider the following simplified
situation: assume that our input space $\Xcal$ only consists of a
finite set of points, that is $\Xcal = \{x_1, ..., x_m\}$ for some
large number $m$. Now consider all possible ways to assign labels to
those data points, that is we consider all possible probability
distributions on $\Xcal$. Given some small set of training points
$(X_i, Y_i)_{i=1,...,n}$ we use some fixed classification rule to
construct a classifier on those points, say the $\knn$ classifier. Now
consider all points of $\Xcal$ which have not been training points and
call this set of points the test set. Of course there exists a label
assignment $P_1$ on the test set for which the classifier makes no
error at all, namely the assignment which has been chosen by the
classifier itself. But there also exists some label assignment $P_2$
on which the classifier makes the largest possible error, namely the
inverse of the assignment constructed by the classifier. In the same
way, we can see that essentially for any given error $R$, we can
construct a probability distribution on $\Xcal$ such that the error of
$f_n$ on the test set has error $R$.  The same reasoning will apply to
any other classifier (for more precise reasoning,
cf. \citealp{HoPep02}).  Thus, averaged over all possible probability
distributions on $\Xcal$, all classifiers $f_n$ will achieve the same
test error: whenever there is a distribution where the classifier performs
well, there is a corresponding ``inverse'' distribution on the test
set, on which the classifier performs poorly. In particular, on
average over all probability distributions, no classifier can be
better than random guessing on the test set! \\

This is a very strong result, and it touches the very base of machine
learning. Does it in fact say that learning is impossible? Well, the
answer is ``it all depends on the underlying assumptions''. The
crucial argument exploited above is that we take the average over {\em
  all} possible probability distributions, and that all probability
distributions are considered to be ``equally likely'' (in the Bayesian
language, we choose a uniform prior over the finite set of
distributions). Those distributions also include cases where the
labels are assigned to points ``without any system''. For example,
somebody could construct a probability distribution over labels by
simply throwing a coin and for each data point, deciding on its true
label by the outcome of the random coin tossing. It seems plausible
that in such a scenario it does not help to know the labels on the
training points --- they are completely independent of the labels of
all other points in the space. In such a
scenario, learning is impossible. \\

The only chance for learning is to exclude such artificial cases. We
need to ensure that there is some inherent mechanism by which we can
use the training labels to generalize successfully to test
labels. Formally, this  means that we have to restrict the
space of probability distributions under consideration. Once we make
such restrictions, the no free lunch theorem breaks down. Restrictions
can come in various forms. For example, we could assume that the
underlying distribution has a ``nice density'' and a ``nice
function $\eta$''. Or we can assume that
there is a distance function on the space and the labels depend in
some ``continuous'' way on the distance, that is points which are
close to each other tend to have similar labels. If we make such
assumptions, it is possible to construct classifiers which exploit
those assumptions (for example, the $\knn$ classifier to exploit
distance structure). And those classifiers will then perform well on
data sets for which the assumptions are satisfied. Of course, the no
free lunch theorem still holds, which means that there will be some
other data sets where this classifier will fail miserably. However,
those will be data sets which come from distributions where the
assumptions are grossly violated. And in those cases it makes complete
sense that a classifier which relies on those assumptions does not
stand a chance any more. \\

The no free lunch theorem is often depicted by a simple figure, see
Figure~\ref{fig-nfl}. The figure shows the performance of two
different classifiers (where, intuitively, the performance of a
classifier is high if it achieves close to the Bayes risk). The $x$-axis
depicts the space of all probability distributions.  Classifier 1
represents a general purpose classifier. It performs moderately on all
kinds of distributions. Classifier 2 depicts a more specialized
classifier which has been tailored towards particular
distributions. It behaves very good on those distributions, but worse
on other distributions. According to the no free lunch theorem, the
average performance, that is the area under the two curves, is the
same for both classifiers. \\

A question which often comes up in the context of no free lunch
theorems is how those theorems fit together with the consistency
theorems proved above. For example, we have seen in
Section~\ref{sec-knn} that the $k$-nearest neighbor classifier is {\em
  universally} consistent, that is it is consistent for {\em any}
underlying probability distribution $P$. Is there a contradiction
between the no free lunch theorem and the consistency statements? The
solution to this apparent paradox lies in the fact that the
consistency statements only treat the limit case of $n \to \infty$. In
the example with the finite data space above, note that as soon as the
sample size is so large that we essentially have sampled each point of
the space at least once, then a classifier which memorizes the
training data will not make any mistake any more. Similar statements
(but a bit more involved) also hold for cases of infinite data
spaces. Thus, no free lunch theorems make statements about some
finite sample size $n$, whereas consistency considers the limit of $n
\to \infty$.  In the finite example
above, note that unless we know the number $m$ of points in the data
space, there is no way we could give any finite sample guarantee on a
classifier. If we have already seen half of the data points, then the
classifier will perform better than if we have only seen 1/100 of all
points. But of course, there is no way we can tell this from a finite
sample. A formal way of stating this is as follows:

\begin{theorem}[Arbitrarily close to random guessing]
Fix some $\eps > 0$. For every $n \in \Nat$ and every classifier $f_n$
there exists a distribution $P$ with Bayes risk 0 such that the
expected risk of $f_n$ is larger than $1/2 - \eps$. 
\end{theorem}

This theorem formally states what we have already hinted above: we can
always construct a distribution such that based on a finite sample
with fixed size, a given classification rule
is not better than random guessing. This and several other versions of
the no free lunch theorem can be found in Section~7 of
\citet{DevGyoLug96}. \\

The no free lunch theorem is one of the most important theorems in
statistical learning. It simply tells us that in order to be able to
learn successfully with guarantees on the behavior of the classifier,
we need to make assumptions on the underlying distribution under
consideration. This fits very nicely to the insights we gained in
Section~\ref{sec-capacity}. There we have seen that in order to
construct consistent classifiers, we need to make assumptions on the
underlying space $\Fcal$ of function one uses. In practice, it makes
sense to combine those statements: first restrict the space of
probability distributions under consideration, and then use a small
function class which is able to model the distributions in this
class. \\

\section{Model based approaches to learning } \label{sec-model-based}

Above we have introduced the standard framework of statistical
learning theory. It has been established as one of the main building
blocks for analyzing machine learning problems and algorithms, but of
course it is not the only approach to do this. In this section we would
like to mention a few other ways to view machine learning problems 
and to analyze them. In particular, we will focus on methods which
deviate from the model-free approach of not making any assumption on
the underlying distribution.

\subsection{The principle of minimum description length}

The classical SLT looks at learning very much from the point of view
of underlying function classes. The basic idea is that good learning
guarantees can be obtained if one uses simple function classes. The
simplicity of a function space can be measured by one out of many
capacity measures such as covering numbers, VC dimension, Rademacher
complexity, and so on. \\

The minimum description length (MDL) approach is based on a different
notion of ''simplicity''. The concept of simplicity used in MDL is
closely related to the literal meaning of simple: an object is called
``simple'' if it can be described by a ``short description'', that is
if one only needs a small number of bits to describe the object. In
the context of MDL, objects can be a sequence of data, or a function,
or a class of functions.  As an example, consider the following
function: ``The function $f:[0,1] \to \{-1, +1\}$ takes the value -1
on the interval $[0,0.3]$ and +1 on $]0.3,1]$.'' For a function of the
class $f:[0,1] \to \{-1, +1\}$, this is a rather compact
description. On the other hand, consider a function $g:[0,1] \to \{-1,
+1\}$ which takes the values +1 and -1 at random positions on
$[0,1]$. In order to describe this functions, the best we
can do is to compile a table with input and output values: \\

\begin{tabular}{l|| l|l|l|l|l}
X    & 0.01 & 0.02 & 0.03 & 0.04 & ... \\
\hline
g(X) & 1    & -1   & -1   & 1    & ... \\
\end{tabular}\\

It is obvious that compared to the code for $f$, the code for $g$ will
be extremely long (even if we ignore the issue that a function on
$[0,1]$ cannot simply be described in a table with countably many
entries). \\

The question is now how we can formulate such a  concept of simplicity in
mathematical terms. A good candidate in this respect is the theory of
data coding and data compression. A naive way to encode a function (on
a finite domain) is to simply provide a table of
input and output values. This can be done with any
function, and is considered to be our baseline code. Now, given a
function $f$, we try to encode it more efficiently. In order to be
able to do this, we need to discover certain ``regularities'' in the
function values. Such a regularity might be that 100 consecutive
entries in the function table have the same value $g(X)$, say +1. Then, instead of
writing +1 in 100 entries  of the table, we can say something
like: ``the following 100 entries have value +1''. This is much shorter
than the naive approach. \\

The MDL framework tries to make use of such insights for learning
purposes. The general intuition is that learning can be interpreted as
finding regularities in the data. If we have to choose between two
functions $f$ and $g$ which have a similar training error, then we
should always prefer the function which can be described by a shorter
code. This is an explicit way to follow Occam's razor, which in machine learning is usually interpreted as ``models should not be more complex than is necessary to
explain the data''.  \\

Omitting all technical details, we would just like to point out one
possible way to turn the MDL idea in a concrete algorithm for
learning. Assume we are given a
function space $\Fcal$, and some training points. One can now try to
pick the function $f \in \Fcal$ which minimizes the following
expression: 

\ba
L(f) + L(\text{Training Data} \condon f)
\ea

Let us explain those two terms. $L(f)$ stands for the length of the code to
encode the function $f$ in the given function class
$\Fcal$. This is some absolute length which is only influenced by the
function itself, and not by the data at hand. We do not go into
details how such a code can be obtained, but the general idea is as
described above: the shorter the code, the simpler the function. One
has to note that in constructing this code, the function space $\Fcal$
might also play a role. For example, one could just encode $f$ by
saying ``take the 23rd function in space $\Fcal$''. In this code, the
length $L(f)$ would just depend on the ordering of the functions in
$\Fcal$. If the function we are looking for occurs early in this
ordering, it has a short code (``function 23''), but if the function occurs
late in this ordering, it has a longer code (``function
4341134''). \\

The term $L(\text{Training Data} \condon f)$ denotes
the length of the code to express the given training data with help of
function $f$. The idea here is simple: if $f$ fits the training
data well, it is trivial to describe the training data. We can simply
use the function $f$ to compute all the labels of the
training points. In this case, the length of the code is very short: 
it just contains the instruction ``apply $f$ to the input to
obtain the correct output''. On the other hand, if the function $f$
does not fit the data so well, it will make some misclassifications on
the training points. To recover the label of the misclassified
training points, we thus need to add some further information to the
code. If, say, training points $X_2$, $X_7$, and $X_{14}$ are
misclassified by $f$, the code for the data might now look as
follows: ``apply $f$ to all input points to
compute the labels. Then flip the labels of training points $X_2$,
$X_7$, and $X_{14}$''. It is clear that the more errors $f$ makes on
the given training points, the
longer this code will get. \\

Intuitively, the two terms $L(f)$ and $L(\text{Training Data} \condon
f)$ play similar roles as some of the quantities in classical SLT. The
term $L(\text{Training Data} \condon f)$ corresponds to the training
error the function $f$ makes on the data. The term $L(f)$ measures the
complexity of the function $f$. In this sense, the sum of both looks
familiar: we sum the training error and some complexity term. One
difference to the classical SLT approach is that the complexity term
is not only computed based on the underlying function class
$\Fcal$, but can depend on the individual function $f$.\\

The approach outline above has often been criticized to be rather
arbitrary: a function that has a short description (small
$L(f)$) under one encoding method may have a long description (large
$L(f)$) under another. How should we decide what description method to
use?
There are various answers to this questions.  The most common one is
the idea of {\em universal coding}. Here codes are associated with
classes of models rather than individual classifiers. There exist
several ways to build universal codes. One of them is as follows. We
decompose the function class ${\cal F}$ into subsets ${\cal F}_1
\subset {\cal F}_2 \subset...$. Then we encode the elements of each
subset with a fixed-length code which assigns each member of ${\cal
  F}_i$ the same code length.  For example, if ${\cal F}_i$ is finite
and has $N$ elements, one may encode each member $f \in \Fcal_i$ using
$\log N$ bits. Or, if the function class $\Fcal_i$ is infinite one can
go over to concepts like the VC dimension to encode the functions in
the class. As in the finite case, each element $f \in \Fcal_i$ will be
encoded with the same length, and this length turns out to be
related to the VC dimension of $\Fcal_i$. Many other ways might be
possible. 
In general, we define the  "coding
complexity of $ {\cal F}_i$" as the smallest uniform code length one can achieve for
encoding the functions $f \in \Fcal_i$. It is uniquely defined and is
independent of and ``universal'' in the sense that it does not rely on
any particular coding scheme. \\

Given a function $f \in \Fcal_i$, we can now go ahead and encode our data in several steps: we 
first encode the index $i$ of the function class, then use the
uniform code described above to encode which element of $\Fcal_i$ the
function $f$ is, and finally code the data with the help of $f$. The
code length then becomes 

\ba
L(i) + L(f  \mid f \in {\cal F}_i) + L( \text{Training Data} | f)
\ea

Note that for the first term $L(i)$ one usually chooses some uniform
code for the integers, which is possible as long as we are dealing
with finitely many function classes $\Fcal_i$.  Then this term is
constant and does not play any role any more.  The middle term $L(f
\mid f \in {\cal F}_i) $ is identical for all $f \in {\cal
  F}_i$. Hence, it does not distinguish between different functions
within the class $\Fcal_i$, but only between functions which come from
different classes $\Fcal_i$. Finally, the last term explains how well
the given function $f$ can explain the given training data. The goal
of this approach is now to choose the
hypothesis $f$ which minimizes this code. \\

In this formulation one can see that the MDL approach is not that far
from standard statistical learning theory approaches. For example, if
we are just given one fixed function class $\Fcal$ (and do not split
it further into smaller sets $\Fcal_i$), the code length essentially
depends on some complexity measure of $\Fcal$ plus a term explaining
how well $f$ fits the given data. In this setting, one can prove
learning bounds which show that MDL learns about as fast as the
classical methods. One can also see that the approach can be closely
related to classical statistical learning approaches based on
compression coefficients \citep{Vapnik95}.
Moreover, in more advanced MDL approaches one may want to assign
different code lengths to different elements of ${\cal F}$. This
approach is then closely related to the PAC-Bayesian approach
(cf. Section \ref{subsec-pac-bayesian}).  Finally, it is worth
mentioning is that under certain assumptions, MDL can be performed in
a consistent way. That is, in the limit of infinitely many data
points, the approach can find the correct model
for the data. \\

Of course there are many details to take care of in the MDL approach,
but this goes beyond the scope of the current paper.
We refer to the monographs \citet{Gruenwald07} and
\citet{Rissanen07} for a comprehensive
treatment of the MDL approach and related concepts.  \\

\subsection{Bayesian methods}

The two traditional schools in statistics are the frequentist and the
Bayesian one. Here we would like to briefly recall their approaches to
inference and discuss their applications to machine learning. Before
we do this, let us introduce the basic setting and some
notation. Traditionally, statistical inference is performed in a
model-based framework. As opposed to the agnostic approach taken in
SLT, we assume that the underlying probability distribution comes from
some particular class of probability distributions $\Pcal$. This class
is usually indexed by one (or several) parameters, that is it has the
form $\Pcal = \{P_{\alpha} \condon \alpha \in A\}$. Here, $A$ is a set
of parameter values, and $P_\alpha$ a distribution. For example, one
could consider the class of normal distributions, indexed by
their means and variances. \\

{\bf The standard frequentist approach to statistics. } Here, the main
goal is to infer, from some given sample of points, the correct
parameters of the underlying distribution. Once the parameter is
known, it is easy to perform tasks such as classification (for example
by using the Bayes classifier corresponding to the estimated
distribution). The most important quantity used in the frequentist
approach to statistics is the {\em likelihood of the parameter} $\alpha$, denoted by
$P( data \condon \alpha)$. This term tells us the probability that,
under the assumption that $\alpha$ is the correct parameter of the
distribution, the given sample is generated. It is used as an
indicator of how good the parameter $\alpha$ ``fits'' the data: if it
is unlikely that the data occurs based on the underlying distribution
$P_{\alpha_1}$, but more likely for $P_{\alpha_2}$, then one would
tend to prefer parameter $\alpha_2$ over $\alpha_1$. This already
hints the way parameters are inferred from the data in the frequentist
setting: the maximum likelihood (ML) approach. Here we choose the
``best'' parameter $\hat\alpha$ by maximizing the data likelihood,
that is
\ba
\hat\alpha(data) = \argmax_{\alpha \in A} P(data \condon \alpha),
\ea
where we have used the notation $\hat\alpha(data)$ to indicate that
$\hat\alpha$ depends on the data.  It is important to note that the
likelihood {\em does not} make any statement about the ``probability
that a parameter $\alpha$ is correct''. All inference about $\alpha$
happens indirectly, by estimating the probability of the data given a
parameter. The same is also true if we want to make confidence
statements. It is impossible to make a statement like ``the
probability that $\alpha$ is correct is larger than something''. First
of all, it is a mathematical problem that we cannot make such a
statement: we simply do not have a probability distribution over
parameters.
If we want to make confidence statements in the traditional approach,
this has to be done in a somewhat peculiar way. A confidence statement
looks as follows: let $I(data) \in A$ be some interval (or more
generally, subset) of parameters which has been constructed from the
data. For example, $I$ could be a symmetric interval of a certain
width around an estimated parameter value $\hat\alpha(data)$, say $I =
[\hat\alpha(data) - c, \hat\alpha(data) + c]$ for some constant
$c$. The set $I$ is called a 95\% confidence interval if
$P_\alpha(\alpha \in I) \geq 95\%$, which is a shorthand for
\ba
P_\alpha\left({data \condon \alpha \in [\hat\alpha(data) - c, \hat\alpha(data) + c]}\right) \geq 95\%.
\ea
Again, it is important to point out that the random quantity in this
statement is $I$, not $\alpha$. The statement only says that if
$\alpha$ happened to be the true parameter, and $I$ is the confidence
set we come up with when looking at the data, then in 95\% of all
samples the true parameter $\alpha$ will lie in $I$. It does not say
that with 95\% probability,
$\alpha$ is the correct parameter! \\

This is one of the reasons why the frequentist approach is sometimes
perceived as unsatisfactory. It only provides a rather indirect way to
perform inference, and confidence statements are hard to grasp. \\

{\bf The Bayesian approach to statistics. } The Bayesian framework is
an elegant way to circumvent some of the problems of the frequentist
one, in particular the indirect mechanism of inference provided by the
ML framework.  From a technical point of view, the main difference
between the Bayesian and the frequentist approach is that the Bayesian
approach introduces some ``prior'' distribution on the parameter
space. That is, we define some distribution $P(\alpha)$ which for each
parameter $\alpha$ encodes how likely we find it that this is a good
parameter to describe our problem. The important point is that this
prior distribution is defined {\em before} we get to see the data
points from which we would like to learn. It should just encode our
past experiences or any other kind of prior knowledge we might
have. Now assume we are given some data points. As in the frequentist
approach, one can compute the likelihood term $P(data \condon
\alpha)$. Combining it with the prior distribution $P(\alpha)$, we can
compute the so-called posterior distribution $P(\alpha \condon
data)$. Up to a normalizing constant independent of $\alpha$, the
posterior is given by the product of the prior and the likelihood
term, that is

\banum \label{eq-posterior}
 P(\alpha \condon data) \propto P(\alpha) P(data  \condon \alpha). 
\eanum

One can say that the posterior distribution arises by ``updating'' the
prior belief using the data we actually have at hand.  As opposed to
the frequentist approach, this posterior is now indeed interpreted as the
probability that $\alpha$ is the correct parameter, given that we have
observed the
data, and the prior distribution is ``correct.''\\

Given the posterior distribution, there are two main principles to do
inference based on it. The first one is to use the
maximum a posteriori (MAP) estimator: 

\banum \label{eq-map}
\hat\alpha = \argmax_{\alpha \in A} P( \alpha \condon data) 
\eanum

Here we come up with one fixed value of $\alpha$, and can now use it
to do the inference we are interested in, in a similar way as we can
use the ML estimator in the frequentist approach. The second, fully Bayesian
way, is to do the inference based on {\em any} parameter $\alpha$, but
then weight the result by the posterior probability that $\alpha$ is
correct. \\

What are advantages and disadvantages of the Bayesian approach? One
advantage is the fact that the Bayesian approach leads to simple,
intuitive statements about the results of a learning algorithm. As
opposed to making complicated confidence statements like the ones in
the standard SLT approach or the traditional frequentist approach to
statistics, in the end one has statements like ``with probability 95\%
we selected the correct parameter $\alpha$''. This comes at a price,
though. The most vehement objection to the Bayesian approach is often
the introduction of the prior itself. The prior does influence our
results, and by selecting different priors one can obtain very
different results on the same data set. %
It is sometimes stated that the influence of the prior is benign and
in the limit, the prior is ``washed out'' by the data, as indicated by
certain consistency results for the Bayesian approach
\citep{Berger85}.  However, this argument is somewhat misleading. On a
finite data set, the whole point of the prior is that it should bias
our inference towards solutions that we consider more likely. So it is
maybe appropriate to say that the Bayesian approach is a convenient
method for \emph{updating} our beliefs about solutions which we had
before taking into account the data.  Among
Bayesian practitioners it is generally accepted that even though
priors are ``wrong'', most of the time they are quite useful in that
Bayesian averaging over parameters leads to good generalization
behavior.  One point in favor of working
with prior distributions is that they are a nice tool to invoke
assumptions on the underlying problem in a rather explicit way.  In
practice, whether or not we should apply Bayesian methods thus depends
on whether we are able to encode our prior knowledge in
the form of a distribution over solutions.\\

On the other hand, SLT is arguably more explicit about model
complexity, an issue which in Bayesian framework is a little harder to
spot. 
As an example for how model complexity is dealt with in the Bayesian
framework, consider the posterior distribution
\eqref{eq-posterior} and the MAP problem \eqref{eq-map}. We 
rewrite this problem by taking the negative logarithm, in which case the
product becomes a sum and the maximization problem becomes a minimization problem:

\ba
\argmin_{\alpha \in A}  \; (\; - \log P(\alpha) - \log P(data \condon \alpha)\; ). 
\ea

(Note that we are a bit sloppy here: applying the logarithm can be
interpreted as a re-pa\-ra\-meter\-ization of the problem. One can show that
MAP solutions are not invariant under such re-parameterizations.) 
Now we are given a sum of two terms which have an interpretation
similar to the approaches we have seen so far.  One could say that the
second term $\log P(data \condon \alpha)$ describes the
model fit to the data, that is it plays a similar role as the training
error $\Remp$ in the standard SLT approach, or the quantity $L(data
\condon f)$ in the minimum description length framework. Continuing
with this analogy, the other term $P(\alpha)$ would then be the term
corresponding to the ``complexity'' of $f$. But how is the prior
probability related to complexity? The general idea is as follows. It
is a simple fact from coding theory that there exist much fewer
``simple'' models than ``complicated'' models. For example, a
``simple'' model class described by one parameter $\alpha \in \{1, 2,
3, ..., 100\}$ contains $100$ models, whereas a slightly less simple
model class using two such parameters already contains $100^2 = 10000$
models. That is, the number of models increases dramatically with the
complexity of the model class. Now assume we want to assign a prior
distribution to the parameters. For finite model classes as above, a
prior distribution simply assigns a certain probability value to each
of the models in the class, and those probability values have to sum
to 1. In case of the first model mentioned above, we need to assign
$100$ numbers to the parameters $\alpha$. If we do the same for the
second model, we have $10000$ probabilities to assign. Now note that
in order to make them sum to 1, the individual prior values per
parameter tends to be much smaller in the second case than in the
first case, simply because the second case requires many more
parameters than the first one. On a higher level, the consequence is
that prior probabilities assigned to elements of ``complex'' model
classes tend to be much lower than prior probabilities for elements of
``simple'' model classes. Hence, then negative logarithm of the priors
of complex function classes is high, whereas the same quantity for
simple model classes is low. All in all, as in the standard SLT
approach, the Bayesian framework implicitly deals with overfitting by
looking at the
trade-off between data fit and model complexity. \\

The literature on general Bayesian statistics is huge.  A classic is
\citet{Cox61}, which introduces the fundamental axioms that allow to
express beliefs using probability calculus. \citet{Jaynes03} puts
those axioms to work and addresses many practical and philosophical
concerns. Another general treatment of Bayesian statistics can be
found in \citet{Ohagan94}. A gentle introduction into Bayesian methods
for machine learning can be found in \citet{Tipping03}, a complete
monograph on machine learning with a strong focus on Bayesian methods
is \citet{Bishop06}. \\

\section{The VC dimension, Popper's dimension, and the number of
  parameters}

We have seen that for statistical learning theory, it is crucial that
one control the capacity of the class of hypothesis from which one
chooses the solution of a learning process. The best known measure of
capacity is a combinatorial quantity termed the VC
dimension. \citet{CorSchVap05} have pointed out that the VC dimension
is related to Popper's notion of the dimension of a theory. Popper's
dimension of a theory is defined as follows (for a discussion of the
terms involved, we refer to \citet{Popper59,CorSchVap05}):
 \begin{quote}
   If there exists, for a theory $t$, a field of singular (but not
   necessarily basic) statements such that, for some number $d$, the
   theory cannot be falsified by any $d$-tuple of the field, although
   it can be falsified by certain $(d+1)$-tuples, then we call $d$ the
   characteristic number of the theory with respect to that field. All
   statements of the field whose degree of composition is less than
   $d$, or equal to $d$, are then compatible with the theory, and
   permitted by it, irrespective of their
   content. \cite[p. 130]{Popper59}
 \end{quote}
 \citet{CorSchVap05} argue that although this definition sounds
 similar to the VC dimension, there is a crucial difference which
 could either be attributed to an error on Popper's side, or to a
 difference between statistical learning and ``active'' learning: in
 Popper's definition, it is enough to find one $(d+1)$-tuple that
 falsifies a theory. E.g, the ``theory'' that two classes of data can
 be separated by a hyperplane could be falsified by three collinear
 points labeled ``$+1$'', ``$-1$'', and ``$+1$''. In the definition
 of the VC dimension, on the other hand, it is required that there
 exists no $(d+1)$-tuple of points which can be shattered, i.e., for
 any $(d+1)$-tuple of points, there exists some labeling such that it
 falsifies the hyperplane, say. The VC dimension of separating
 hyperplanes in $\R^n$ is $n+1$, while the Popper dimension of
 separating hyperplanes is always $2$, independent of $n$. Whilst one
 could fix this by adding some noise to the points in the Popper
 dimension, thus ruling out the existence of non-generic
 configurations, it may be more interesting to relate the difference
 between the two definitions to the fact that for Popper, the
 scientist trying to perform induction is actively looking for points,
 or experiments, that might be able to falsify the current hypothesis,
 while Vapnik and Chervonenkis devised their capacity measure to
 characterize the generalization error of a learning procedure where
 the incoming measurement points are
 generated randomly according to a certain probability distribution.\\

 Interestingly, Popper also discusses the link between his capacity
 measure and the number of a parameters in an algebraic
 characterization of a hypothesis class \citep{CorSchVap05}, stating
 that ``[...] the number of freely determinable parameters of a set of
 curves by which a theory is represented is characteristic for the
 degree of falsifiability [...]'' (cited after
 \citet{CorSchVap05}). From the point of view of statistical learning
 theory, Popper here falls into the same trap hole as classical
 statistics sometimes does. In learning theory, it is not the number
 of parameters but the capacity which determines the generalization
 ability. Whilst the number of parameters sometimes coincides with,
 e.g., the VC dimension (the above mentioned hyperplanes being an
 example), there are also important cases where this is not true. For
 instance, it has been pointed out that the class of thresholded sine
 waves on $\R$, parameterized by a single real frequency parameter,
 has infinite VC dimension \citep{Vapnik95}.\\

 We conclude with \citet{CorSchVap05} that Popper has at an
 astonishingly early point in time identified some of the crucial
 aspects of what constitutes the capacity of a class of hypothesis. If
 he had had the full developments of statistical learning theory at
 his disposal, he might have been able to utilize them to address
 certain other shortcomings of his approach, in particular by using
 bounds of statistical learning theory to make statements about
 certain notions of reliability or generalization ability of theories.

\section{Conclusion}
At first glance, methods for machine learning are impressive in that
they automatically extract certain types of ``knowledge'' from
empirical data. The above description, however, has shown that in
fact, none of this knowledge is created from scratch.\\

In the Bayesian
view of machine learning, the data only serves to update one's prior
--- we start with a probability distribution over hypothesis, and end
of up with a somewhat different distribution that reflects what we
have seen in between. For a subjective Bayesian, learning is thus
nothing but an update of one's beliefs which is consistent with the
rules of probability theory. Statements regarding how well the
inferred solution works are generally not made, nor are they necessary
--- for an orthodox Bayesian.\\

In the framework of statistical learning theory, on the other hand, we
start with a class of hypotheses, and use the empirical data to select
one hypothesis from the class. One can show that if the data
generating mechanism is benign, then we can assert that the difference
between the training error and test error of a hypothesis from the
class is small. ``Benign'' here can take different guises; typically
it refers to the fact that there is a stationary probability law that
independently generates all individual observations, however other
assumptions (e.g., on properties of the law) can also be
incorporated. The class of hypothesis plays a role analogous to the
prior, however, it does not need to reflect one's beliefs. Rather, the
statements that we obtain are conditional on that class in the sense
that if the class is bad (in the sense that the ``true'' function
cannot be approximated within the class, or in the sense that there is
no ``true'' function, e.g., the data is completely random) then the
result of our learning procedure will be unsatisfactory in that the
upper bounds on the test error will be too large. Typically, either
the training error will be too large, or the confidence term,
depending on the capacity of the function class, will be too
large. %
It is
appealing, however, that statistical learning theory generally avoids
metaphysical statements about aspects of the ``true'' underlying
dependency, and thus is precisely by referring to the difference
between training and test error.\\

While the above are the two main theoretical schools of machine
learning, there are other variants some of which we have briefly
mentioned in this article. Importantly, none of them get away without
making assumptions, and learning is never a process that starts from a
tabula rasa and automatically generates knowledge.

\section*{Acknowledgments}

We would like to thank Wil Braynen, David Corfield, Peter Gruenwald, Joaquin
Quinonero Candela, Ingo Steinwart, and Bob Williamson
for helpful comments on the manuscript.

\bibliography{general_bib,bibfile}

\end{document}